%% file: main.tex
\documentclass[10pt,twocolumn,letterpaper]{article}

\usepackage[pagenumbers]{cvpr} 
\usepackage{multirow}
\input{preamble}

\definecolor{cvprblue}{rgb}{0.21,0.49,0.74}
\usepackage[pagebackref,breaklinks,colorlinks,allcolors=cvprblue]{hyperref}

\title{InfiniDepth: Arbitrary-Resolution and Fine-Grained Depth Estimation \\ with Neural Implicit Fields}

\author{
Hao Yu$^{1,2*}$~~~
Haotong Lin$^{1*}$~~~
Jiawei Wang$^{1*}$~~~
Jiaxin Li$^{1}$~~~
Yida Wang$^{2}$~~~
Xueyang Zhang$^{2}$~~~\\
Yue Wang$^{1}$~~~
Xiaowei Zhou$^{1}$~~~
Ruizhen Hu$^{3}$~~~
Sida Peng$^{1\dag}$~~~
\\
\small{
$^{1}$Zhejiang University \ 
$^{2}$Li Auto \
$^{3}$Shenzhen University \
}
\\
\small{
\textit{Project Page:} \href{https://zju3dv.github.io/InfiniDepth/}{zju3dv.github.io/InfiniDepth}
}
}

\begin{document}
% \maketitle
\twocolumn[\maketitle\vspace{0em}\input{figures/intro/teaser.tex}\bigbreak]
\setcounter{footnote}{0}
\renewcommand{\thefootnote}{\fnsymbol{footnote}}
\footnotetext[1]{Equal contribution. \dag{Corresponding} author.}
\renewcommand{\thefootnote}{\arabic{footnote}}

\input{sec/0_abstract}    
\input{sec/1_intro}

\input{sec/2_related_work}
\input{sec/3_method}
\input{sec/4_experiments}
\input{sec/5_conclusion}

{
    \small
    \bibliographystyle{ieeenat_fullname}
    \bibliography{main}
}

\input{sec/X_suppl.tex}

\end{document}

%% file: preamble.tex
\usepackage{booktabs}  % for professional-looking thick rules
\usepackage{colortbl}  % for coloring table cells
\definecolor{lightpink}{RGB}{233, 23, 115}
\definecolor{first}{RGB}{220, 245, 220}   % softer light green for 1st place
\definecolor{second}{RGB}{255, 255, 200}  % light yellow for 2nd place
\definecolor{third}{RGB}{200, 230, 255}   % light blue for 3rd place
\definecolor{lightblue}{RGB}{100, 149, 237} 
\definecolor{standardblue}{RGB}{0, 112, 192}
\definecolor{lightgreen}{RGB}{144,238,144}

%% file: figures/intro/teaser.tex
\begin{center}
    \vspace{-8mm}
    \captionsetup{type=figure}
    \includegraphics[width=0.98\textwidth]{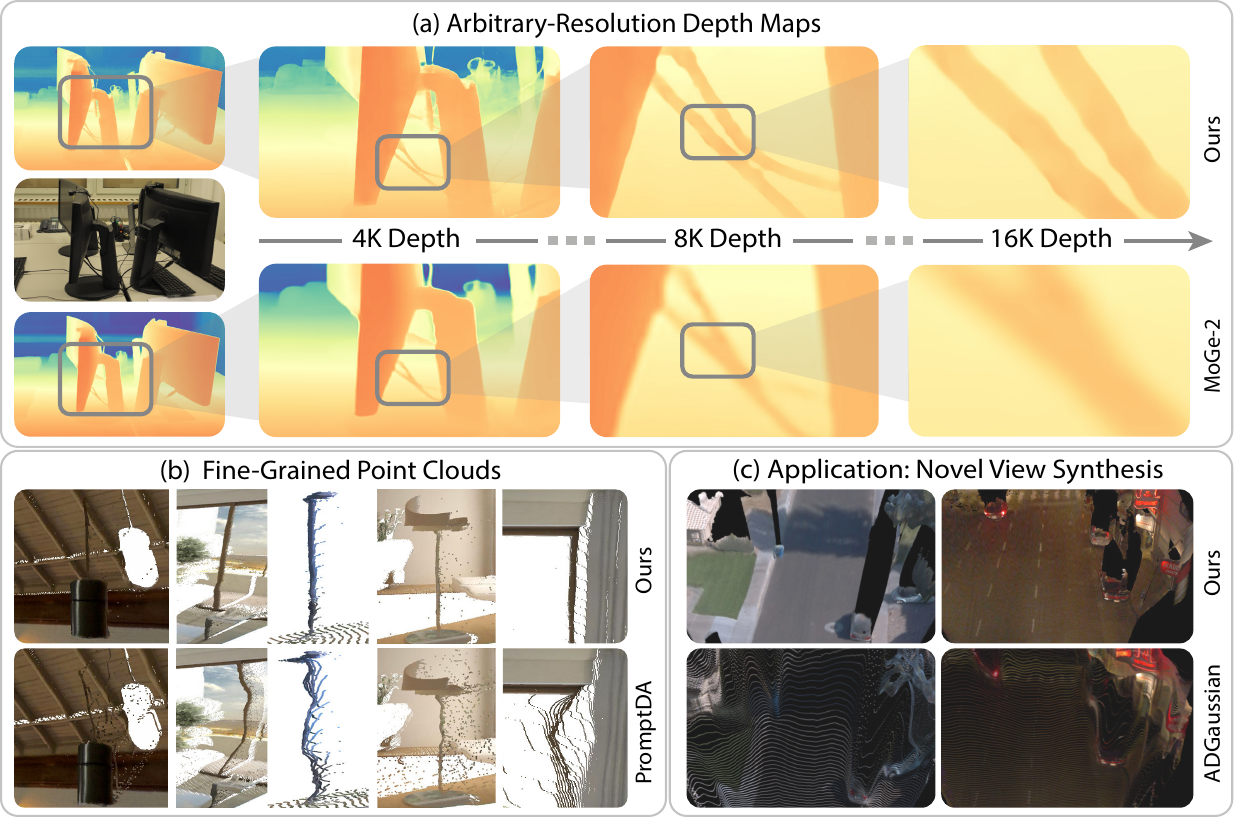}
    \captionof{figure}{\textbf{InfiniDepth} is a new depth representation which models depth as neural implicit fields, enabling arbitrary-resolution and fine-grained depth estimation. It also benefits novel view synthesis under large viewpoint shifts with fewer holes and artifacts.
     }
    \vspace{1mm}
    \label{fig:teaser}
\end{center}

%% file: sec/0_abstract.tex
\begin{abstract}
Existing depth estimation methods are fundamentally limited to predicting depth on discrete image grids. Such representations restrict their scalability to arbitrary output resolutions and hinder the geometric detail recovery.
This paper introduces \textbf{InfiniDepth}, which represents depth as neural implicit fields.
Through a simple yet effective local implicit decoder, we can query depth at continuous 2D coordinates, enabling arbitrary-resolution and fine-grained depth estimation.
To better assess our method's capabilities, we curate a high-quality 4K synthetic benchmark from five different games, spanning diverse scenes with rich geometric and appearance details.
Extensive experiments demonstrate that InfiniDepth achieves state-of-the-art performance on both synthetic and real-world benchmarks across relative and metric depth estimation tasks, particularly excelling in fine-detail regions.
It also benefits the task of novel view synthesis under large viewpoint shifts, producing high-quality results with fewer holes and artifacts.
\end{abstract}

%% file: sec/1_intro.tex
\section{Introduction}
\label{sec:intro}
Monocular depth estimation (MDE) is a fundamental task in computer vision, with widespread applications in autonomous driving and robotics.
Some traditional methods~\cite{liu2015learning, liu2015deep} represent the depth map as a graph-structured output with conditional random fields (CRFs), showing some success in the early stages but are limited in scalability and detail prediction due to optimization complexity.

With the development of deep learning, mainstream depth estimation methods~\cite{ranftl2021vision, yang2024depth, yang2024depth2, bochkovskii2024depth, wang2025moge, ke2024repurposing,xu2025pixel} adopt regular 2D grids to represent depth maps, as this representation is naturally compatible with modern neural network architectures.
Although these methods demonstrate strong generalization, they struggle to produce high-resolution depth maps while preserving fine details, and tend to fail to accurately predict depth in regions with significant geometric variations.
Fundamentally, these limitations stem from the discrete grid-based depth representation, 
which constrains depth prediction at fixed grid locations, inherently limiting output resolution to the training image size.
Moreover, predicting depth on entire grids requires either convolutional upsampling or linear projection from latents to depth patches.
The former introduces smoothing effects, while the latter struggles to capture local geometric variations—both sacrificing high-frequency details.

In this paper, we present \textbf{InfiniDepth}, a new depth representation that models depth as neural implicit fields, enabling arbitrary-resolution and fine-grained depth estimation.
Specifically, an input image is encoded by a vision transformer into multi-stage feature tokens, followed by a reassemble block that constructs a feature pyramid.
Then, for any continuous 2D coordinate $(x,y)$, we gather spatially aligned features from the pyramid within a local window and feed them into a lightweight MLP to predict depth.
Unlike prior methods constrained to grid-based depth prediction, InfiniDepth adopts a continuous and localized prediction paradigm.
It is no longer constrained by training resolutions and naturally produces arbitrary-resolution depth maps with fine details (Fig.~\ref{fig:teaser} (a)).
The localized prediction further excels at capturing geometric variations, producing fine-grained point clouds (Fig.~\ref{fig:teaser} (b)).

Another benefit of InfiniDepth is its ability to enhance novel view synthesis (NVS) quality under large viewpoint shifts.
Specifically, recent feed-forward NVS methods~\cite{xu2025depthsplat, song2025adgaussian} predict pixel-aligned depth and Gaussian parameters. 
Unprojecting such a discrete per-pixel depth map produces a surface point cloud with strong density imbalance due to perspective projection and surface orientation, thereby degrading NVS quality under large viewpoint shifts.
To address this limitation, we design a depth query strategy that allocates sub-pixel query budgets proportionally to each pixel's corresponding 3D surface element, producing spatially uniform 3D points on object surfaces.
With the uniform 3D points, our method produces high-quality novel view synthesis, with markedly fewer holes and reduced artifacts under large viewpoint shifts (Fig.~\ref{fig:teaser} (c)).

To better assess resolution scalability and detail prediction capabilities, we curate Synth4K, a high-quality benchmark collected from five different games, covering diverse scenes with 4K ground-truth depth maps.
We also construct high-frequency depth masks to isolate fine-detail regions for targeted evaluation of detail prediction.
Extensive experiments on Synth4K and real-world benchmarks demonstrate that InfiniDepth consistently achieves state-of-the-art performance across both relative and metric depth estimation tasks, with particularly strong results in fine-detail regions.
Furthermore, we demonstrate that InfiniDepth combined with a depth query strategy can benefit novel view synthesis under large viewpoint shifts.

In summary, this work has the following contributions:
\begin{itemize}
  \item We propose a new depth representation that models depth as neural implicit fields and demonstrate its capability for arbitrary-resolution and fine-grained depth estimation.
  \item We design a depth query strategy that produces uniformly distributed 3D points on object surfaces, improving novel view synthesis quality under large viewpoint shifts.
  \item We curate Synth4K, a high-quality 4K benchmark for evaluating depth estimation methods at high resolution and fine geometric details.
\end{itemize}

%% file: sec/2_related_work.tex
\section{Related Work}
\label{sec:related_work}
\begin{figure*}[ht]
    \centering
    \includegraphics[width=\linewidth]{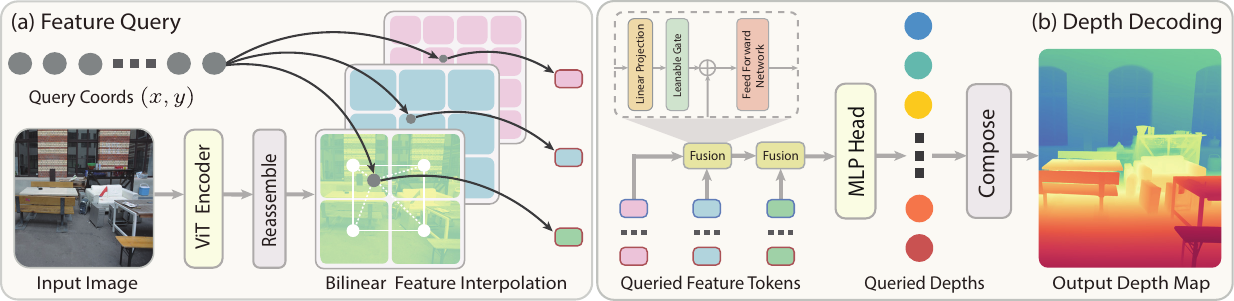}
    \caption{
        \textbf{Overview of InfiniDepth.} (a) Feature Query: given an input image and a continuous query 2D coordinate, we extract feature tokens from multiple layers of the ViT encoder, and query local features for the coordinate at each scale through bilinear interpolation.
        (b) Depth Decoding: given the multi-scale local features queried at the continuous coordinate, we hierarchically fuse features from high spatial resolution to low spatial resolution, and decode the fused feature to the depth value through a MLP head.
      } 
    \label{fig:pipeline}
    \vspace{-3mm}
\end{figure*}
\paragraph{Relative Depth Estimation.}
Relative depth estimation aims to infer a normalized depth map without absolute scale. 
Recent works~\cite{ranftl2021vision, yang2024depth,yang2024depth2, wang2025moge} adopt Vision Transformer (ViT) backbones~\cite{dosovitskiy2020image} with convolutional decoders to regress 2D discretized depth maps. 
DepthAnything~\cite{yang2024depth, yang2024depth2} improves generalization by combining labeled and large-scale unlabeled data, while MoGe~\cite{wang2025moge} enhances geometric accuracy with affine-invariant point maps and optimal training supervision.
Diffusion-based methods~\cite{ke2024repurposing, xu2025pixel, he2024lotus, gui2025depthfm} model the distribution of depth maps, with Marigold~\cite{ke2024repurposing} leveraging pretrained diffusion priors and PPD~\cite{xu2025pixel} refining depth boundaries via a semantics-prompted DiT.
However, all these methods represent depth as discrete 2D grids, limiting resolution scalability and fine detail recovery.
\vspace{-2mm}
\paragraph{Metric Depth Estimation.} 
Early metric depth methods~\cite{bhat2021adabins, fu2018deep, bhat2023zoedepth, li2024patchfusion} typically formulate the problem as a global distribution classification task or fine-tune depth models on datasets with metric depth annotations.
Recent approaches~\cite{yin2021learning, hu2024metric3d, yin2023metric3d, piccinelli2024unidepth, bochkovskii2024depth} address the ambiguity by incorporating camera intrinsics, while others~\cite{liu2024depthlab, wang2025depth, zuo2025omni, viola2025marigold, lin2025prompting} leverage sparse depth as additional inputs to improve accuracy.
For example, PriorDA~\cite{wang2025depth} and Omni-DC~\cite{zuo2025omni} complete depth maps under various patterns of sparse depth inputs, while Marigold-DC~\cite{viola2025marigold} parameterizes the scale and shift of metric depth and optimizes them iteratively.
PromptDA~\cite{lin2025prompting} introduces a novel depth prompt module for accurate estimation, but fine-grained geometry recovery remains challenging.
In this work, we demonstrate that InfiniDepth combined with sparse depth inputs can significantly enhance metric depth estimation, especially in predicting fine-grained geometry details.
\vspace{-2mm}
\paragraph{Implicit Neural Representations.}
\emph{Implicit Neural Representations} (INRs) map continuous coordinates to signals and have been widely applied in 3D reconstruction and beyond.
NeRF~\cite{mildenhall2021nerf} models scenes as neural radiance fields and PiFU~\cite{saito2019pifu} uses a pixel-aligned implicit function to relate image pixels to 3D human geometry.
The paradigm has also been extended to 2D images, optical flow, and multi-view scene representation.
LIIF~\cite{chen2019learning} learns continuous image representation with an implicit function and AnyFlow~\cite{jung2023anyflow} achieves arbitrary scale optical flow with implicit representation.
DeFiNe~\cite{guizilini2022depth} proposes an implicit multi-view scene representation, but architectural constraints limit it to low-resolution outputs.
Inspired by these advances, we represent depth as neural implicit fields along with a simple yet effective implicit decoder, enabling arbitrary-resolution and fine-grained depth estimation.

%% file: sec/3_method.tex
\section{Method}

\label{sec:method}
% Overview
Given a single RGB image, our goal is to estimate depth for any continuous 2D coordinate in the image plane. The overview of our method is illustrated in Fig.~\ref{fig:pipeline}.

% Section 3.1: Preliminary: Neural Implicit Field
\subsection{Representing Depth as Neural Implicit Fields}
\label{sec:Continuous Depth Representation}
Neural implicit fields model signals $\mathbf{y}$ as an implicit function of the continuous coordinates $\mathbf{x}$ parameterized by a neural network:
\begin{equation}
    \mathbf{y} = F_{\theta}(\mathbf{x}),
    \label{eq:nif}
\end{equation}
where $F_{\theta}$ is typically implemented as a multi-layer perceptron (MLP).
Compared to explicit representations such as voxels or image grids whose fidelity scales with discretization, neural implicit field models fine-grained geometry in a resolution-agnostic manner with fewer parameters.

We extend the concept of neural implicit fields to represent depth, which models depth estimation as an implicit function that maps any continuous 2D coordinate $(x,y) \in [0, W] \times [0, H]$ to depth value $d_I(x,y)$, conditioned on the input RGB image $I \in \mathbb{R}^{H \times W \times 3}$:
\begin{equation}
    d_I(x,y) = N_{\theta}(I,(x,y)),
    \label{eq:depth-nif}
\end{equation}
where $N_{\theta}$ is parameterized by a neural network.

% Section 3.2: Multi-Scale Local Implicit Decoder
\subsection{Multi-Scale Local Implicit Decoder}
\label{subsec:implicit-decoder}
We instantiate $N_{\theta}$ in Eq.~\ref{eq:depth-nif} as a multi-scale local implicit decoder, which reassembles and aggregates features from multiple layers of the image encoder for any continuous query coordinate $(x,y)$, with a lightweight MLP head to predict depth values.
This simple yet effective decoder consists of two modules: \textbf{Feature Query} and \textbf{Depth Decoding}.
\paragraph{Feature Query.}
The input image $I$ is firstly encoded by a Vision Transformer to obtain a set of feature tokens.
Following~\cite{ranftl2021vision}, we design a reassemble block which extracts feature tokens from multiple ViT layers and projects them to different hidden dimensions.
To capture fine local details and preserve global semantics, we upsample shallow-layer features (\textcolor{pink}{pink} and \textcolor{lightblue}{blue} tokens in Fig.~\ref{fig:pipeline} (a)) to higher spatial resolutions, while retaining deeper-layer features (\textcolor{lightgreen}{green} tokens) at their native resolution.
In this way, we construct a feature pyramid $\{f^{k}\}_{k=1}^{L}$ with $f^{k} \in \mathbb{R}^{h_k \times w_k \times C^{k}}$.

\begin{figure}[ht]
    \centering
    \includegraphics[width=\linewidth]{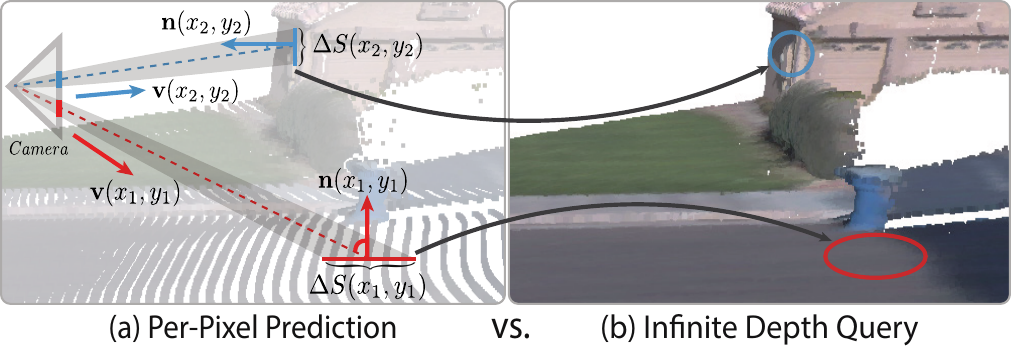}
    \caption{
        \textbf{Advantage of Our Infinite Depth Query.} The \textcolor{standardblue}{blue} and \textcolor{red}{red} highlighted regions represent areas with different depths, surface normals, and viewing directions.
        Per-pixel depth prediction leads to strong density imbalance in these regions due to perspective projection and surface orientation, while Infinite Depth Query applies sub-pixel query with adaptive weights to generate uniformly distributed point clouds.
      } 
    \label{fig:sampling}
    \vspace{-3mm}
\end{figure} 

For a continuous query coordinate $(x, y) \in [0, W] \times [0, H]$ in the image plane, we first map it to the corresponding location $(x_k, y_k) \in [0, w_k] \times [0, h_k]$ at the $k$-th feature-pyramid scale:
\[
(x_k, y_k) = \Bigl(x \cdot \frac{w_k}{W},\; y \cdot \frac{h_k}{H}\Bigr).
\]
For each scale $k$, we define the local grid neighborhood around $(x_k, y_k)$ as as $\mathcal{N}_k(x_k,y_k)$, which is $\bigl\{ (i,j)\;\big|\; i \in \{\lfloor x_k \rfloor,\, \lfloor x_k \rfloor + 1\},\; j \in \{\lfloor y_k \rfloor,\, \lfloor y_k \rfloor + 1\} \bigr\}$,
and aggregate features from this neighborhood using bilinear interpolation, yielding a feature token
$f^{k}_{(x,y)} \in \mathbb{R}^{1 \times C^{k}}$ for the query coordinate $(x, y)$ at scale $k$.

\paragraph{Depth Decoding.} 
Given the multi-scale local descriptors $\{f^{k}_{(x,y)}\}_{k=1}^{L}$, we fuse them hierarchically from shallow (detail) to deep (semantic) features, aiming to better capture fine-grained geometric variations, preserve both local details and global context, and achieve high-precision and robust depth decoding.

Let $\mathbf{h}_1 := f^{1}_{(x,y)} \in \mathbb{R}^{C_1}$ denote the queried feature at the shallowest (highest-resolution) scale. For each scale $k=1,\dots,L-1$, we hierarchically fuse $\mathbf{h}_k$ with the next-scale feature $f^{k+1}_{(x,y)} \in \mathbb{R}^{C_{k+1}}$ using a residual gated fusion block:
\begin{equation}
    \mathbf{h}_{k+1} = \mathrm{FFN}_k\Big( f^{k+1}_{(x,y)} + \mathbf{g}_k \odot \mathrm{Linear}(\mathbf{h}_k) \Big),
\end{equation}
where $\mathrm{Linear}(\cdot)$ denotes a linear projection to match the feature dimension, $\mathbf{g}_k \in (0,1)^{C_{k+1}}$ is a learnable channel-wise gate, and $\odot$ denotes element-wise multiplication.
Here, $\mathrm{FFN}_k(\cdot)$ denotes a two-layer feed-forward network with non-linear activation. This process is repeated from $k=1$ to $L-1$, resulting in the final fused feature $\mathbf{h}_L \in \mathbb{R}^{C_L}$ at the deepest scale.
Finally, the depth value at $(x, y)$ is predicted by an MLP head:
\begin{equation}
    d_I(x, y) = \mathrm{MLP}(\mathbf{h}_L).
\end{equation}

\begin{figure}[ht]
    \centering
    \includegraphics[width=\linewidth]{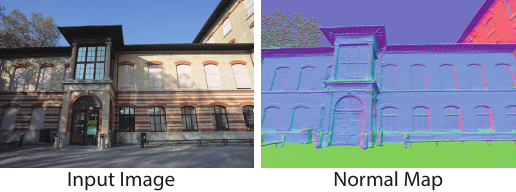}
    \caption{
        \textbf{Normal map from implicit fields through \textit{torch autograd},} indicating high-quality internal geometry of our model.
        }
        \label{fig:normal}
    \vspace{-3mm}
\end{figure} 

% Section 3.3: Sampling 3D Uniform Geometry
\subsection{Infinite Depth Query}
\label{subsec:point-sampling}
Unprojecting a discrete per-pixel depth map produces a surface point cloud with strong density imbalance due to perspective projection and surface orientation (Fig.~\ref{fig:sampling} (a)), thereby degrading NVS quality under large viewpoint shifts.
We present a depth query strategy that generates approximately uniform 3D points on visible surfaces by leveraging our implicit depth field.

The key insight is that the 3D surface area $\Delta S(x,y)$ corresponding to each pixel depends on two geometric factors: 
(1) \textbf{depth-squared scaling}—pixels at greater depth cover a surface area that grows quadratically with distance ($\propto d^2$), and 
(2) \textbf{surface orientation effect}—when the surface normal deviates from the viewing direction, its projection onto the image is compressed, causing each pixel to cover a larger actual surface area.
We counteract these effects by allocating sub-pixel query budgets proportionally to each pixel's corresponding 3D surface element.

Specifically, we first query depth at pixel coordinates $(x,y)$ and back-project them to 3D points $\mathbf{X}(x,y)$, then derive an adaptive weight $w(x,y)$ that estimates the differential surface area $\Delta S(x,y)$ at each pixel location:
\begin{equation}
    w(x,y) = \frac{d_I(x,y)^{2}}{\bigl|\mathbf{n}(x,y) \cdot \mathbf{v}(x,y)\bigr| + \varepsilon} \;\propto\; \Delta S(x,y),
\end{equation}
where $d_I(x,y)$ denotes the queried depth, $\mathbf{n}(x,y)$ is the surface normal, $\mathbf{v}(x,y)$ represents the unit viewing direction, and $\varepsilon$ is a small constant for numerical stability.
In this formulation, $d_I(x,y)^2$ accounts for depth-squared scaling, while $|\mathbf{n}(x,y) \cdot \mathbf{v}(x,y)|$ compensates for surface orientation effect, together approximating the 3D surface area subtended by each pixel.
$\mathbf{n}(x,y)$ is computed from the Jacobian of $\mathbf{X}(x,y)$ with respect to continuous image coordinates, leveraging the differentiable nature of our implicit depth field (Fig.~\ref{fig:normal}):
\begin{equation}
\mathbf{n}(x,y) =
\frac{\partial_x \mathbf{X}(x,y) \times \partial_y \mathbf{X}(x,y)}
     {\bigl\|\partial_x \mathbf{X}(x,y) \times \partial_y \mathbf{X}(x,y)\bigr\|} \in \mathbb{R}^{3}.
\end{equation}

Based on $w(x,y)$, we allocate adaptive query budgets and uniformly distribute sub-pixel query coordinates within each pixel patch.
Querying $d_I(x,y)$ at these continuous coordinates and back-projecting to 3D yields a point cloud with approximately uniform surface coverage (Fig.~\ref{fig:sampling} (b)).

See \textit{supp.} for implementation details and visualizations.

% Section 3.4: Implementation Details
\subsection{Implementation Details}
\label{subsec:Details}
\paragraph{Network Architecture.} We adopt the DINOv3~\cite{simeoni2025dinov3} ViT-Large model as our image encoder.
We extract feature maps from layers 4, 11, and 23 of the encoder and project them to hidden dimensions of 256, 512, and 1024, respectively.
The feature maps from layers 4 and 11 are then upsampled by factors of 4 and 2, respectively.
See \textit{supp.} for more details on the network design, as well as an evaluation of computational efficiency and parameter count.
\paragraph{Training Data and Strategies.} 
Given our goal of achieving fine-grained depth estimation, we exclusively train our model on synthetic datasets, as real-world datasets often contain noisy and incomplete depth maps.
We utilize Hypersim~\cite{roberts2021hypersim}, VKITTI~\cite{cabon2020virtual}, TartanAir~\cite{wang2020tartanair}, IRS~\cite{wang2019irs}, etc., along with several high-resolution datasets including UnrealStereo4K~\cite{Tosi2021CVPR} and UrbanSyn~\cite{gomez2025all}.

Due to the properties of our depth representation, we can flexibly supervise sparse samples instead of the entire depth map. Specifically, we randomly draw $N$ coordinate-depth pairs and compute the $l1$ loss over these points:
\begin{equation}
    \mathcal{L} = \frac{1}{N} \sum_{i=1}^{N} |d_i - \hat{d}_i|,
\end{equation}
where $d_i$ is the ground truth, and $\hat{d}_i$ is the predicted depth.
We train our model using the AdamW optimizer with a learning rate of $1 \times 10^{-5}$. It's trained for 800k steps using 8 NVIDIA GPUs, with a batch size of 4 per GPU.
See \textit{supp.} for more details on training data and strategies.

%% file: sec/4_experiments.tex
\section{Experiments}
\label{sec:experiments}
% 4.1 Synth4K
\input{tables/synth4k.tex}
\input{tables/synth4k_lidar.tex}
\input{tables/realdata.tex}
\input{tables/realdata_lidar.tex}
\subsection{Synth4K}
\label{subsec:synth4k}
To assess zero-shot generalization, prior methods are commonly evaluated on real-world benchmarks.
However, ground-truth depth in these datasets is typically low-resolution and sparse, often failing to capture fine geometric structures such as edges and high-frequency details.
This makes it challenging to reliably evaluate models on high-resolution and fine-grained depth estimation.

To address this limitation, we curate a high-quality synthetic benchmark named Synth4K, specifically designed for zero-shot evaluation.
Synth4K consists of RGB-D data collected from five different games (denoted as Synth4K-1 to Synth4K-5).
Each subset contains hundreds of 4K-resolution image pairs, spanning diverse indoor and outdoor environments.
We further compute a multi-scale Laplacian energy map for each depth image and sample pixels proportionally to the energy to construct a binary high-frequency (HF) mask.
This mask highlights geometrically detailed regions and enables targeted evaluation.
See \textit{supp.} for more implementation details and visualizations of Synth4K.

Compared with existing benchmarks, Synth4K provides significantly higher depth-map resolution and substantially improved detail coverage, establishing a stronger benchmark for high-resolution and fine-grained depth estimation.

% 4.2 Experiment Setup
\subsection{Experimental Setup}
\label{subsec:eval}
% \paragraph{Training datasets.} 
\paragraph{Relative Depth Estimation.} Following prior work, we evaluate zero-shot relative depth estimation on five real-world datasets: KITTI~\cite{geiger2013vision}, ETH3D~\cite{schops2017multi}, NYUv2~\cite{silberman2012indoor}, ScanNet~\cite{dai2017scannet}, and DIODE~\cite{vasiljevic2019diode}.
The $\delta_1$ accuracy is reported in Table~\ref{tab:realdata_depth_comparison}.
To demonstrate our method's capability for arbitrary-resolution and fine-grained depth estimation, we further conduct extensive experiments on Synth4K, including evaluating depth predictions over the full 4K-resolution images, as well as assessing fine-detail prediction performance in HF-masked regions.
We report $\delta_{0.5}$, $\delta_1$, and $\delta_2$ in Table~\ref{tab:synth_depth_comparison}.
$\delta_{0.5}$, $\delta_1$, and $\delta_2$ denote the percentage of pixels satisfying $\max\left(\frac{d}{d^*}, \frac{d^*}{d}\right) < 1.25^{0.5}$, $1.25^{1}$, and $1.25^{2}$, respectively, where $d$ is the predicted depth and $d^*$ is the ground-truth depth.

\paragraph{Metric Depth Estimation.} To demonstrate that InfiniDepth is also effective for metric depth estimation, we incorporate sparse depth inputs using the depth prompt module proposed in~\cite{lin2025prompting}, referred to as \textbf{Ours-Metric} for clarity.
We report $\delta_{0.01}$, $\delta_{0.02}$, $\delta_{0.04}$ accuracy in Table \ref{tab:synth_depth_comparison_lidar} and Table \ref{tab:realdata_depth_comparison_lidar}.
$\delta_{0.01}$, $\delta_{0.02}$, $\delta_{0.04}$ denote the percentage of pixels where $\max\left(\frac{d}{d^*}, \frac{d^*}{d}\right) < 1.01$, $1.02$, $1.04$, respectively.
These stricter thresholds are adopted because metric depth estimation with sparse depth inputs typically achieves higher prediction accuracy and therefore warrants more stringent evaluation criteria.

\begin{figure*}[ht]
    \centering
    \includegraphics[width=\linewidth]{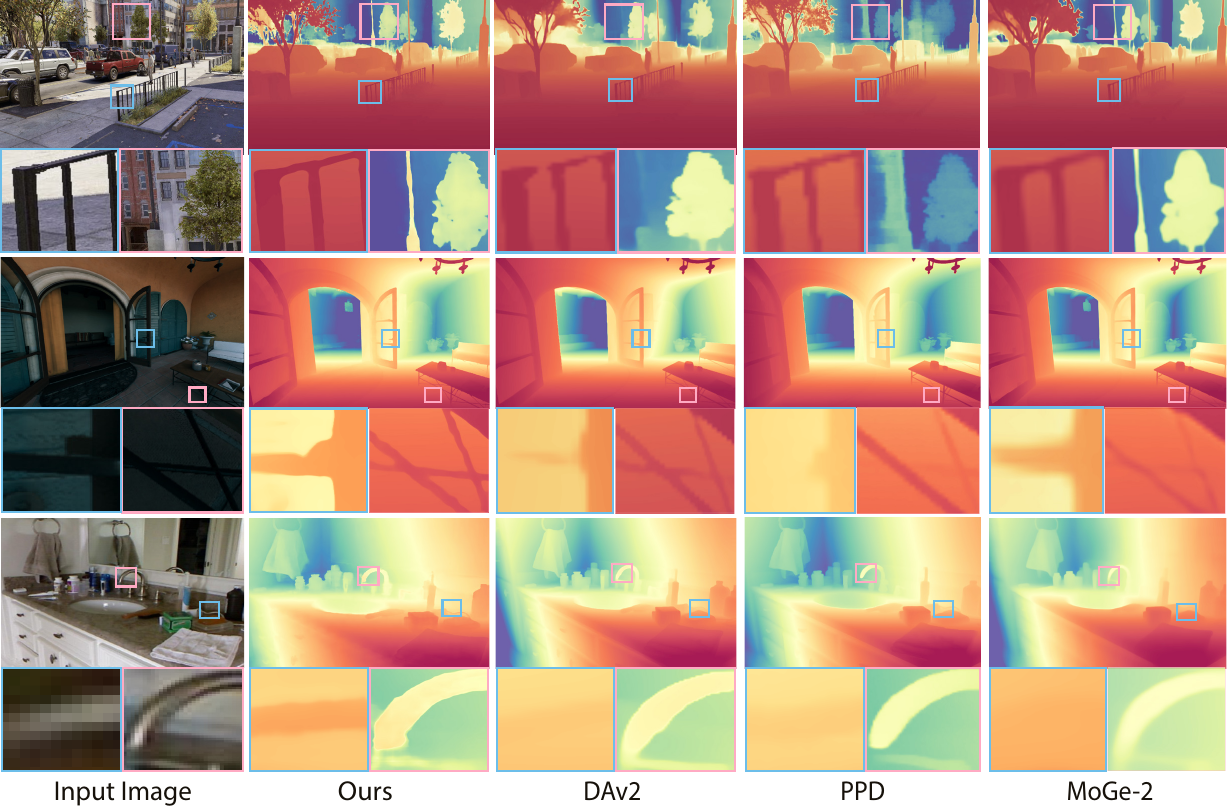}
    \caption{
        \textbf{Qualitative comparisons for relative depth estimation.} The first two rows show prediction results on Synth4K, while the last row shows real-world data with low resolution RGB inputs. The boxes highlight detail regions upsampled to higher resolution for baselines, while our method directly predicts at this resolution.
        More comparisons can be found in the \textit{supp.}.
    }
    \label{fig:mono_compare}
    \vspace{-3mm}
\end{figure*}

\begin{figure*}[ht]
    \centering
    \includegraphics[width=\linewidth]{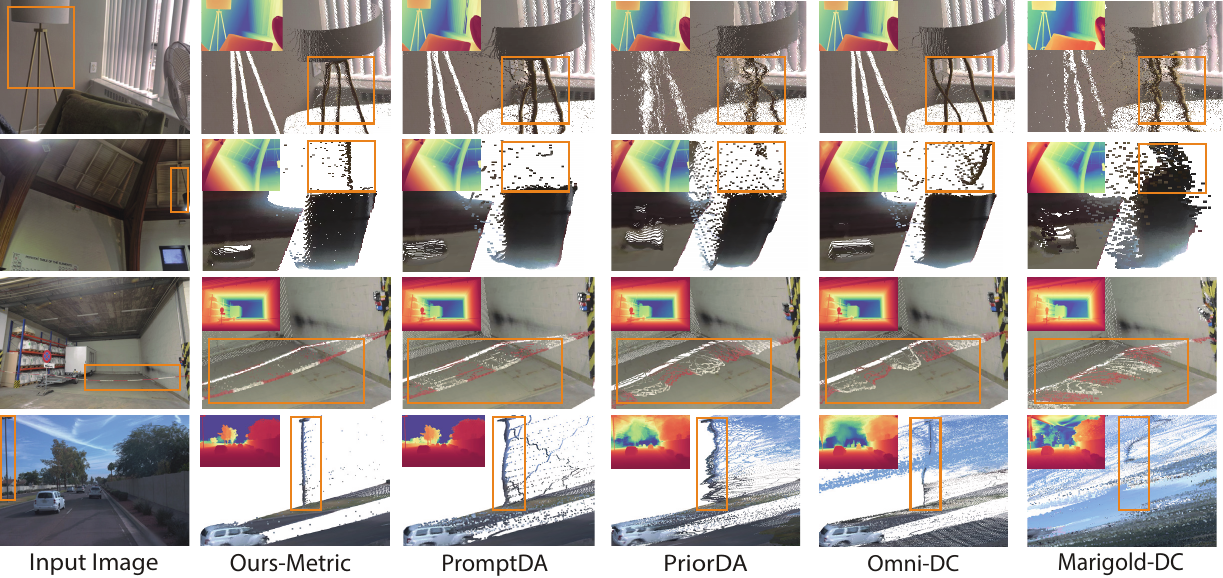}
    \caption{
        \textbf{Qualitative comparisons for metric depth estimation.} The boxes highlight the high-frequency geometric details. 
      } 
    \label{fig:lidar_compare}
    \vspace{-3mm}
\end{figure*}

\input{tables/ablation.tex}

% \subsection{Comparisons with the State of the Art.}
\subsection{Comparisons with the State of the Art}
We compare our approach to two categories of SOTA methods on both Synth4K and real-world benchmarks:
(1) relative depth estimation using only RGB inputs, and
(2) metric depth estimation with additional sparse depth.
For relative depth estimation, we evaluate against DepthAnything~\cite{yang2024depth}, DepthAnythingV2~\cite{yang2024depth2}, DepthPro~\cite{bochkovskii2024depth}, MoGe~\cite{wang2025moge}, MoGe2~\cite{wang2025moge2}, Marigold~\cite{viola2025marigold}, and PPD~\cite{xu2025pixel}, aligning predictions to ground-truth depth before evaluation.
For metric depth estimation, we compare with depth completion approaches, including Marigold-DC~\cite{viola2025marigold}, Omni-DC~\cite{zuo2025omni}, PriorDA~\cite{wang2025depth}, and PromptDA~\cite{lin2025prompting}.
To ensure fair comparisons, we use the same input resolution across all baselines and the same sparse depth samples for metric methods.
On Synth4K, baseline outputs are upsampled to 4K, whereas InfiniDepth is queried directly at 4K.

As shown in Table \ref{tab:synth_depth_comparison} and Table \ref{tab:synth_depth_comparison_lidar}, our method significantly outperforms all existing methods across all metrics on Synth4K, highlighting its strength in high-resolution and fine-grained depth estimation.
On real-world benchmarks, \textbf{Ours} performs on par with current SOTA methods, while \textbf{Ours-Metric} achieves clear improvements over existing metric depth estimation methods, as shown in Table \ref{tab:realdata_depth_comparison} and Table \ref{tab:realdata_depth_comparison_lidar}.
(Marigold-DC suffers from VAE-based quantization loss, as discussed in~\cite{xu2025pixel}, leading to low metric accuracy.)
Qualitative depth map (Fig.~\ref{fig:mono_compare}) and point cloud comparisons further illustrate the advantages of our approach in producing accurate and detailed predictions.
\begin{figure}[ht]
    \vspace{-2mm}
    \centering
    \includegraphics[width=\linewidth]{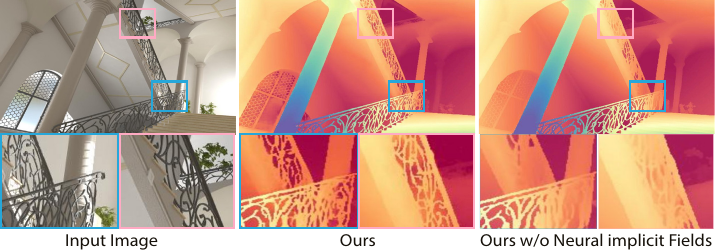}
    \caption{
       \textbf{Qualitative ablations on depth representation.} The boxes highlight the fine-detail regions.
      } 
    \label{fig:ablation depth_representation}
    \vspace{-2mm}
\end{figure}
\subsection{Ablations and Analysis}
\label{subsec:ablation}
\paragraph{Depth Representation.} To verify the effectiveness of our depth representation, we compare it with a baseline that predicts depth on discrete grids using a DPT decoder~\cite{ranftl2021vision}.
Both models share the same encoder (DINOv3 ViT-Large) and training data (Hypersim).
The quantitative results show that representing depth as neural implicit fields yields substantially better performance for metric depth estimation (Table~\ref{tab:ablation}), with moderate gains for relative depth estimation (See \textit{supp.}).
This gap is expected. Sparse depth inputs greatly reduce the ambiguity of metric depth estimation, allowing more convincing and consistent results.
Our depth representation—together with its localized prediction mechanism—further enhances depth precision, yielding clear improvements in both quantitative metrics and visual quality.
In contrast, RGB-only relative depth estimation suffers from high depth ambiguity, causing quantitative metrics to saturate. Nevertheless, our representation consistently recovers finer geometric details, as shown in Fig.~\ref{fig:ablation depth_representation}.

\paragraph{Design Choices for Implicit Decoder.}
We ablate the multi-scale feature query and fusion mechanism in our implicit decoder, against a baseline that samples features only from the single-scale final feature map of the image encoder for each query coordinate.
The quantitative ablation results in Table \ref{tab:ablation} demonstrate that multi-scale feature query mechanism brings significant improvements across datasets.
We also compare more detailed design choices in \textit{supp.}, including (1) explicitly learning offsets between query coordinates and grid neighborhood vs. bilinear interpolation, (2) employing a cross-attention mechanism for feature querying at given coordinates vs. a shared MLP, etc.
\paragraph{Image Encoder.} We investigate the impact of different image encoders in our framework by comparing DINOv3~\cite{simeoni2025dinov3} and DINOv2~\cite{oquab2023dinov2}, both using ViT-Large backbones.
Quantitative results are summarized in Table~\ref{tab:ablation}.

See \textit{supp.} for more ablations and analysis of our method.

\subsection{Application: Single-View Novel View Synthesis} 
\label{subsec:nvs}
\begin{figure}[ht]
    \vspace{-3mm}
    \centering
    \includegraphics[width=\linewidth]{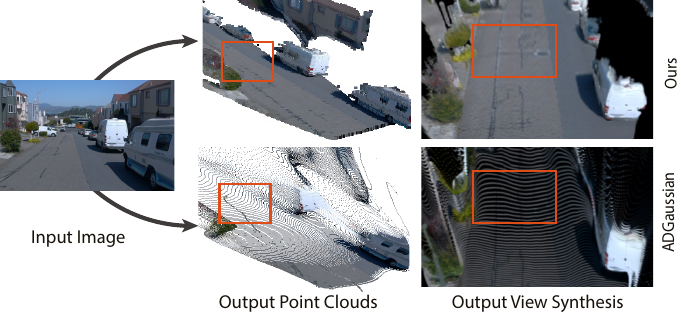}
    \caption{
       \textbf{NVS driven by InfiniDepth and ADGaussian~\cite{song2025adgaussian}.} The boxes highlight regions with geometric holes in the ADGaussian predictions. Refer to \textit{supp.} for more results.
      } 
    \label{fig:nvs}
    \vspace{-2mm}
\end{figure}
We demonstrate that InfiniDepth combined with the depth query strategy (Sec.~\ref{subsec:point-sampling}) significantly improves single-view novel view synthesis (NVS) under large viewpoint shifts.

Specifically, we extend our depth model with a lightweight Gaussian Splatting (GS) head. See \textit{supp.} for more details on the GS head design and training.
At inference time, we first apply the depth query strategy to generate uniformly distributed points on visible surfaces, which serve as Gaussian centers, then feed them to the GS head to predict Gaussian attributes and render novel views under large viewpoint shifts.
As shown in Fig.~\ref{fig:teaser} (c) and Fig.~\ref{fig:nvs}, ADGaussian~\cite{song2025adgaussian}, which predicts pixel-aligned depth, often exhibits noticeable geometric holes and artifacts. In contrast, InfiniDepth produces more complete and stable novel view synthesis results even under large viewpoint shifts.

%% file: tables/synth4k.tex
\begin{table*}[t]
    \centering
    \renewcommand{\arraystretch}{1.0}
    \scalebox{0.8}{%
    \begin{tabular}{c|c|ccc|ccc|ccc|ccc|ccc}
    \specialrule{1.2pt}{0pt}{0pt}
    \multirow{2}{*}{Type} & \multirow{2}{*}{Method} & \multicolumn{3}{c|}{Synth4K-1} & \multicolumn{3}{c|}{Synth4K-2} & \multicolumn{3}{c|}{Synth4K-3} & \multicolumn{3}{c|}{Synth4K-4} & \multicolumn{3}{c}{Synth4K-5} \\
    \cline{3-5} \cline{6-8} \cline{9-11} \cline{12-14} \cline{15-17}
    & & $\delta_{0.5}$ & $\delta_1$ & $\delta_2$ & $\delta_{0.5}$ & $\delta_1$ & $\delta_2$ & $\delta_{0.5}$ & $\delta_1$ & $\delta_2$ & $\delta_{0.5}$ & $\delta_1$ & $\delta_2$ & $\delta_{0.5}$ & $\delta_1$ & $\delta_2$ \\
    \specialrule{1.2pt}{0pt}{0pt}
    \multirow{7}{*}{\begin{tabular}{c}Full\\Image\end{tabular}} 
    & DepthAnything~\cite{yang2024depth} & \cellcolor{second}70.4 & \cellcolor{third}83.8 & \cellcolor{third}93.0	& \cellcolor{second}77.9 & \cellcolor{second}88.2 & \cellcolor{third}95.2 & \cellcolor{second}77.4 & \cellcolor{second}88.6 & \cellcolor{second}96.0	& 83.9 & 92.8 & 96.6 & \cellcolor{second}84.3 & \cellcolor{second}93.0 & 97.4 \\
    & DepthAnythingv2~\cite{yang2024depth2} & 67.3 & 81.3 & 91.0 & \cellcolor{third}76.0 & \cellcolor{third}88.1 & \cellcolor{second}95.4 & \cellcolor{third}71.4 & 85.5 & \cellcolor{third}95.3 & 86.1 & 94.1 & \cellcolor{third}97.4 & 78.6 & 92.1 & 97.6 \\
    & DepthPro~\cite{bochkovskii2024depth} & 63.5 & 80.2 & 91.2	& 66.7 & 83.1 & 93.5 	& 61.2 & 80.2 & 92.1	& 87.1 & 94.1 & 97.1 	& 73.9 & 89.1 & 96.7 \\
    & MoGe~\cite{wang2025moge} & \cellcolor{third}69.3 & 83.7 & 92.7 & 72.8 & 86.2 & 94.1 & 70.6 & \cellcolor{third}85.6 & 94.0 & \cellcolor{third}89.2 & \cellcolor{third}94.6 & 97.0 & \cellcolor{third}81.1 & \cellcolor{third}92.7 & \cellcolor{third}97.7 \\  
    & MoGe-2~\cite{wang2025moge2} & 69.0 & \cellcolor{second}84.2 & \cellcolor{second}93.4 & 73.5 & 86.6 & 94.3 & 70.9 & 85.3 & 94.0 & \cellcolor{first}90.4 & \cellcolor{second}95.3 & \cellcolor{second}97.6 & 80.7 & 92.4 & \cellcolor{second}97.9 \\
    & Marigold~\cite{viola2025marigold} & 54.6 & 72.9 & 85.1 & 57.2 & 75.6 & 87.8 & 55.6 & 73.7 & 85.7 & 79.3 & 90.7 & 95.6 & 66.5 & 84.5 & 93.3 \\  
    & PPD~\cite{xu2025pixel} & 61.5 & 81.1 & 92.5 & 62.2 & 84.6 & 93.9 & 57.5 & 82.8 & 93.9 & 85.6 & 94.1 & 97.0 & 69.1 & 90.4 & 96.5 \\  
    & \textbf{Ours} & \cellcolor{first}74.3 & \cellcolor{first}89.0 & \cellcolor{first}96.1 & \cellcolor{first}80.4 & \cellcolor{first}92.2 & \cellcolor{first}97.0 & \cellcolor{first}82.0 & \cellcolor{first}93.9 & \cellcolor{first}97.8 & \cellcolor{second}89.7 & \cellcolor{first}95.5 & \cellcolor{first}98.0 & \cellcolor{first}88.5 & \cellcolor{first}96.3 & \cellcolor{first}98.8 \\
    \specialrule{1.2pt}{0pt}{0pt}
    \multirow{7}{*}{\begin{tabular}{c}High-Freq\\Details\end{tabular}} 
    & DepthAnything~\cite{yang2024depth} & 43.4 & 61.3 & 78.3 & 41.0 & 59.4 & 77.4 & 44.3 & 62.1 & \cellcolor{third}80.2 & 55.1 & 70.3 & 82.0 & 53.1 & 70.8 & 86.0 \\
    & DepthAnythingv2~\cite{yang2024depth2} & 43.0 & 60.6 & 77.9 & 41.4 & 60.1 & 78.2 & 41.8 & 60.7 & 80.0 & 59.3 & 73.9 & 84.7 & 49.2 & 70.3 & 86.6 \\
    & DepthPro~\cite{bochkovskii2024depth} & 43.4 & 62.4 & 80.6 & 38.4 & 58.8 & \cellcolor{third}79.3 & 38.2 & 58.6 & 79.6 & 62.6 & \cellcolor{third}76.1 & \cellcolor{third}85.3 & 53.3 & 73.1 & \cellcolor{third}89.0 \\
    & MoGe~\cite{wang2025moge} & \cellcolor{third}48.8 & \cellcolor{third}65.8 & \cellcolor{third}80.9 & \cellcolor{third}43.9 & \cellcolor{third}61.6 & 77.9 & \cellcolor{third}45.9 & \cellcolor{third}62.9 & 79.4 & \cellcolor{third}64.4 & 75.7 & 83.9 & \cellcolor{third}60.6 & \cellcolor{third}76.2 & 88.5 \\  
    & MoGe-2~\cite{wang2025moge2} & \cellcolor{second}48.9 & \cellcolor{second}66.5 & \cellcolor{second}82.6 & \cellcolor{second}44.3 & \cellcolor{second}62.5 & \cellcolor{second}79.3 & \cellcolor{second}46.0 & \cellcolor{second}63.4 & \cellcolor{second}80.6 & \cellcolor{first}66.7 & \cellcolor{second}78.2 & \cellcolor{second}85.8 & \cellcolor{second}61.4 & \cellcolor{second}77.3 & \cellcolor{second}89.4 \\
    & Marigold~\cite{viola2025marigold} & 35.8 & 54.0 & 72.0 & 30.8 & 49.4 & 69.9 & 33.4 & 51.4 & 71.1 & 54.2 & 69.9 & 81.2 & 43.9 & 63.2 & 81.1 \\  
    & PPD~\cite{xu2025pixel} & 42.3 & 61.6 & 79.6 & 36.6 & 58.3 & 77.8 & 36.9 & 58.5 & 78.0 & 61.6 & 75.3 & 84.4 & 48.3 & 70.1 & 86.3 \\  
    & \textbf{Ours} & \cellcolor{first}49.2 & \cellcolor{first}67.5 & \cellcolor{first}83.1 & \cellcolor{first}46.7 & \cellcolor{first}65.6 & \cellcolor{first}81.9 & \cellcolor{first}52.5 & \cellcolor{first}69.0 & \cellcolor{first}83.1 & \cellcolor{second}65.3& \cellcolor{first}78.2 & \cellcolor{first}87.3 & \cellcolor{first}63.9 & \cellcolor{first}79.5 & \cellcolor{first}90.7 \\
    \specialrule{1.2pt}{0pt}{0pt}
    \end{tabular}
    } % end of resizebox
    \vspace{-1mm}
    \caption{\textbf{Zero-shot relative depth estimation on Synth4K.} The top-3 results are highlighted as \colorbox{first}{~first~}, \colorbox{second}{~second~}, and \colorbox{third}{~third~}.}
    \label{tab:synth_depth_comparison}
\end{table*}

%% file: tables/synth4k_lidar.tex
\begin{table*}[t]
    \centering
    \renewcommand{\arraystretch}{1.1}
    \scalebox{0.79}{%
    \begin{tabular}{c|c|ccc|ccc|ccc|ccc|ccc}
    \specialrule{1.2pt}{0pt}{0pt}
    \multirow{2}{*}{Type} & \multirow{2}{*}{Method} & \multicolumn{3}{c|}{Synth4K-1} & \multicolumn{3}{c|}{Synth4K-2} & \multicolumn{3}{c|}{Synth4K-3} & \multicolumn{3}{c|}{Synth4K-4} & \multicolumn{3}{c}{Synth4K-5} \\
    \cline{3-5} \cline{6-8} \cline{9-11} \cline{12-14} \cline{15-17}
    & & ${\delta_{0.01}}$ & ${\delta_{0.02}}$ & ${\delta_{0.04}}$ & ${\delta_{0.01}}$ & ${\delta_{0.02}}$&  ${\delta_{0.04}}$ & ${\delta_{0.01}}$ & ${\delta_{0.02}}$&  ${\delta_{0.04}}$ & ${\delta_{0.01}}$ & ${\delta_{0.02}}$&  ${\delta_{0.04}}$ & ${\delta_{0.01}}$ & ${\delta_{0.02}}$&  ${\delta_{0.04}}$ \\
    \specialrule{1.2pt}{0pt}{0pt}
    \multirow{5}{*}{\begin{tabular}{c}Full\\Image\end{tabular}} 
    & Marigold-DC~\cite{viola2025marigold}  & 19.5 & 31.9 & 48.0	&13.2 & 22.4 & 36.1	&18.6 & 32.1 & 49.5	&26.9 & 40.5 & 54.1	&18.0 & 31.4 & 49.0 \\
    & Omni-DC~\cite{zuo2025omni} & 38.8 & 46.0 & 54.1 & 38.4 & 43.8 & 52.5 & 44.0 & 49.5 & 55.1 & 37.9 & 43.2 & 58.9 & 43.4 & 50.5 & 55.7 \\
    & PriorDA~\cite{wang2025depth} & \cellcolor{third}44.8 & \cellcolor{third}67.2 & \cellcolor{third}80.7 & \cellcolor{third}47.3 & \cellcolor{third}67.9 & \cellcolor{third}78.6 & \cellcolor{third}55.5 & \cellcolor{third}75.4 & \cellcolor{third}85.0 & \cellcolor{third}61.9 & \cellcolor{third}78.4 & \cellcolor{third}88.0 & \cellcolor{third}54.0 & \cellcolor{third}75.9 & \cellcolor{third}86.9 \\
    & PromptDA~\cite{lin2025prompting} & \cellcolor{second}65.0 & \cellcolor{second}79.8 & \cellcolor{second}88.0 & \cellcolor{second}66.3 & \cellcolor{second}78.1 & \cellcolor{second}85.4 & \cellcolor{second}72.0 & \cellcolor{second}84.8 & \cellcolor{second}90.8 & \cellcolor{second}78.8 & \cellcolor{second}88.6 & \cellcolor{second}93.1 & \cellcolor{second}69.2 & \cellcolor{second}84.8 & \cellcolor{second}91.2  \\
    & \textbf{Ours-Metric} & \cellcolor{first}78.0 & \cellcolor{first}86.7 & \cellcolor{first}92.0 & \cellcolor{first}76.6 & \cellcolor{first}83.6 & \cellcolor{first}89.0 & \cellcolor{first}83.8 & \cellcolor{first}90.1 & \cellcolor{first}93.5 & \cellcolor{first}87.2 & \cellcolor{first}92.0 & \cellcolor{first}95.0 & \cellcolor{first}83.1 & \cellcolor{first}89.8 & \cellcolor{first}93.5 \\
    \specialrule{1.2pt}{0pt}{0pt}
    \multirow{5}{*}{\begin{tabular}{c}High-Freq\\Details\end{tabular}} 
    & Marigold-DC~\cite{viola2025marigold} & 9.0 & 15.8 & 26.0 & 5.3 & 9.7 & 17.2 & 8.3 & 15.0 & 24.9 & 13.4 & 22.5 & 34.1 & 10.3 & 18.8 & 31.8 \\
    & Omni-DC~\cite{zuo2025omni} & \cellcolor{third}18.4 & \cellcolor{third}26.4 & \cellcolor{third}36.3 & \cellcolor{third}12.4 & \cellcolor{third}19.0 & \cellcolor{third}28.4 & \cellcolor{third}22.9 & \cellcolor{third}30.3 & \cellcolor{third}37.9 & \cellcolor{third}21.8 & 30.1 & 42.1 & \cellcolor{third}24.1 & \cellcolor{third}34.1 & 44.7 \\
    & PriorDA~\cite{wang2025depth} & 12.6 & 21.7 & 33.7 & 8.5 & 15.1 & 24.9 & 13.4 & 21.7 & 31.5 & 20.2 & \cellcolor{third}31.9 & \cellcolor{third}45.8 & 19.0 & 31.8 & \cellcolor{third}45.6 \\
    & PromptDA~\cite{lin2025prompting} & \cellcolor{second}21.1 & \cellcolor{second}33.1 & \cellcolor{second}45.7 & \cellcolor{second}15.3 & \cellcolor{second}24.5 & \cellcolor{second}36.6 & \cellcolor{second}24.7 & \cellcolor{second}35.3 & \cellcolor{second}45.3 & \cellcolor{second}32.0 & \cellcolor{second}45.2 & \cellcolor{second}57.2 & \cellcolor{second}27.3 & \cellcolor{second}41.4 & \cellcolor{second}54.0\\
    & \textbf{Ours-Metric} & \cellcolor{first}33.2 & \cellcolor{first}46.5 & \cellcolor{first}58.7 & \cellcolor{first}24.0 & \cellcolor{first}34.9 & \cellcolor{first}47.8 & \cellcolor{first}37.2 & \cellcolor{first}47.6 & \cellcolor{first}56.5 & \cellcolor{first}45.5 & \cellcolor{first}57.5 & \cellcolor{first}68.2 & \cellcolor{first}38.8 & \cellcolor{first}52.0 & \cellcolor{first}63.5 \\
    \specialrule{1.2pt}{0pt}{0pt}
    \end{tabular}
    } % end of resizebox
    \vspace{-1mm}
    \caption{\textbf{Zero-shot metric depth estimation on Synth4K.} The top-3 results are highlighted as \colorbox{first}{~first~}, \colorbox{second}{~second~}, and \colorbox{third}{~third~}.}
    \label{tab:synth_depth_comparison_lidar}
\end{table*}

%% file: tables/realdata.tex
\begin{table}[t]
    \centering
    \renewcommand{\arraystretch}{1.1}
    \scalebox{0.97}{%
    \resizebox{\columnwidth}{!}{%
    \begin{tabular}{c| c |c| c| c| c}
    \specialrule{1.2pt}{0pt}{0pt}
    \multirow{2}{*}{Method} & KITTI & ETH3D & NYUv2 & ScanNet & DIODE \\
    \cline{2-2} \cline{3-3} \cline{4-4} \cline{5-5} \cline{6-6}
    & $\delta_1$ & $\delta_1$ & $\delta_1$ & $\delta_1$ & $\delta_1$ \\
    \hline
    DepthAnything~\cite{yang2024depth} & 97.5 & 98.4 & \cellcolor{third}97.8 & 97.8 & 97.3 \\
    DepthAnythingV2~\cite{yang2024depth2} & 96.7 & 97.8 & 97.3 & 97.4 & 97.0 \\
    DepthPro~\cite{bochkovskii2024depth} & 97.5 & 98.0 & 97.6 & \cellcolor{third}97.9 & 97.1 \\
    MoGe~\cite{wang2025moge} & \cellcolor{second}98.3 & \cellcolor{third}98.9 & \cellcolor{second}98.0 & \cellcolor{second}98.2 & \cellcolor{third}97.4 \\
    MoGe-2~\cite{wang2025moge2} & \cellcolor{first}98.3 & \cellcolor{second}99.0 & \cellcolor{first}98.2 & \cellcolor{first}98.4 & \cellcolor{second}97.4 \\
    Marigold~\cite{viola2025marigold} & 94.2 & 96.8 & 95.8 & 93.9 & 94.7 \\
    PPD~\cite{xu2025pixel} & 97.3 & 98.3 & 97.2 & 97.3 & 96.2 \\
    \textbf{Ours-Relative} & \cellcolor{third}97.9 & \cellcolor{first}99.1 & 97.6 & 97.3 & \cellcolor{first}97.4 \\
    \specialrule{1.2pt}{0pt}{0pt}
    \end{tabular}
    }
    }% end of resizebox
    % \vspace{-1mm}
    \caption{\textbf{Zero-shot relative depth estimation on real-world datasets.} The top-3 results are highlighted.}
    \label{tab:realdata_depth_comparison}
\end{table}

%% file: tables/realdata_lidar.tex
\begin{table}[t]
    \centering
    \renewcommand{\arraystretch}{1.15}
    \scalebox{0.97}{%
    \resizebox{\columnwidth}{!}{%
    \begin{tabular}{c| c |c| c| c| c}
    \specialrule{1.2pt}{0pt}{0pt}
    \multirow{2}{*}{Method} & KITTI & ETH3D & NYUv2 & ScanNet & DIODE \\
    \cline{2-2} \cline{3-3} \cline{4-4} \cline{5-5} \cline{6-6}
    & ${\delta_{0.01}}$ & ${\delta_{0.01}}$ & ${\delta_{0.01}}$  & ${\delta_{0.01}}$ & ${\delta_{0.01}}$  \\
    \hline
    Marigold-DC~\cite{viola2025marigold} & 36.6 & 72.6 & 71.4 & 76.7 & 84.2 \\ 
    Omni-DC~\cite{zuo2025omni}  & 17.5 & 57.4 & 62.1 & 55.8 & 83.3 \\
    PriorDA~\cite{wang2025depth}  & \cellcolor{third}54.1 & \cellcolor{third}85.7 & \cellcolor{third}78.1 & \cellcolor{third}82.7 & \cellcolor{third}94.3 \\
    PromptDA~\cite{lin2025prompting} & \cellcolor{second}58.3  & \cellcolor{second}92.8 & \cellcolor{second}83.6 & \cellcolor{second}87.0 & \cellcolor{second}97.3 \\
    \textbf{Ours-Metric} & \cellcolor{first}63.9 & \cellcolor{first}96.7 & \cellcolor{first}86.9 & \cellcolor{first}90.4 & \cellcolor{first}98.4 \\
    \specialrule{1.2pt}{0pt}{0pt}
    \end{tabular}
    } % end of resizebox
    } % end of scalebox
    % \vspace{-1mm}
    \caption{\textbf{Zero-shot metric depth estimation on real-world datasets.} The top-3 results are highlighted.}
    \label{tab:realdata_depth_comparison_lidar}
\end{table}

%% file: tables/ablation.tex
\begin{table*}[t]
    \centering
    \renewcommand{\arraystretch}{1.2}
    \scalebox{2.03}{%
    \resizebox{\columnwidth}{!}{%
    \begin{tabular}{c|c|c|c|c|c|c|c|c|c|c}
    \specialrule{1.2pt}{0pt}{0pt}
    \multirow{2}{*}{Ablation} & Synth4K-1 & Synth4K-2 & Synth4K-3 & Synth4K-4 & Synth4K-5 & KITTI & ETH3D & NYUv2 & ScanNet & DIODE  \\
    \cline{2-11}
    & $\delta_{0.01}$ & $\delta_{0.01}$ & $\delta_{0.01}$ & $\delta_{0.01}$ & $\delta_{0.01}$ & $\delta_{0.01}$ & $\delta_{0.01}$ & $\delta_{0.01}$ & $\delta_{0.01}$ & $\delta_{0.01}$ \\
    \specialrule{1.2pt}{0pt}{0pt}
    \textbf{Full Model} & \textbf{72.7} & \textbf{73.5} & \textbf{78.2} & \textbf{81.5} & \textbf{79.4} & \textbf{61.7} & \textbf{93.9} & \textbf{84.7} & \textbf{88.5} & \textbf{97.6}  \\
    w/o Neural Implicit Fields  & 62.4 & 65.1 & 66.5 & 73.2 & 68.9 & 49.0 & 88.9 & 81.2 & 84.2 & 95.4 \\
    w/o Multi-Scale Query & \underline{66.6} & \underline{67.4} & \underline{70.8} & 77.0 & \underline{72.4} & \underline{59.7} & 88.7 & \underline{82.5} & \underline{86.2} & 95.6 \\
    w/o DINOv3~\cite{simeoni2025dinov3} & 63.8 & 66.2 & 67.9 & \underline{77.0} & 71.7 & 57.9 & \underline{90.1} & 80.8 & 83.2 & \underline{95.8}  \\
    % w/o Both & 42.2 & 79.6 & 72.0 & 71.3 & 90.1 & 56.5 & 58.4 & 58.1 & 69.2 & 62.3 \\
    \specialrule{1.2pt}{0pt}{0pt}
    \end{tabular}
    }
    }% end of resizebox
    \caption{\textbf{Quantitative ablations on different datasets.} The \textbf{best} and  \underline{second best} are highlighted. See Sec.~\ref{subsec:ablation} for details.}
    \label{tab:ablation}
\end{table*}

%  

%% file: sec/5_conclusion.tex
\section{Conclusion and Discussions}
\label{sec:conclusion}
This paper presents a new depth representation that models depth as neural implicit fields. This formulation enables depth estimation at arbitrary continuous 2D coordinates while better preserving fine-grained geometric details. The effectiveness of the proposed representation is validated on both Synth4K and real-world benchmarks, across different tasks including relative and metric depth estimation.
Our method achieves significant improvements in depth prediction accuracy and detail recovery. Combined with a depth query strategy, it further benefits single-view novel view synthesis under large viewpoint shifts.

\paragraph{Limitations and Future Work.} This work focuses on monocular depth estimation and is trained only on single-view depth data, so when applied to videos it does not explicitly enforce temporal consistency and may exhibit flickering across frames. Future work includes extending our depth representation to multi-view settings to improve temporal stability and 3D consistency. We hope that InfiniDepth will inspire further research on high-resolution, fine-grained depth estimation and its integration into broader 3D perception and reconstruction pipelines.

%% file: sec/X_suppl.tex
\clearpage
\appendix
\setcounter{page}{1}
\maketitlesupplementary
% \section{Rationale}
% \label{sec:rationale}
% % 
% Having the supplementary compiled together with the main paper means that:
% % 
% \begin{itemize}
% \item The supplementary can back-reference sections of the main paper, for example, we can refer to \cref{sec:intro};
% \item The main paper can forward reference sub-sections within the supplementary explicitly (e.g. referring to a particular experiment); 
% \item When submitted to arXiv, the supplementary will already included at the end of the paper.
% \end{itemize}
% % 
% To split the supplementary pages from the main paper, you can use \href{https://support.apple.com/en-ca/guide/preview/prvw11793/mac#:~:text=Delete%20a%20page%20from%20a,or%20choose%20Edit%20%3E%20Delete).}{Preview (on macOS)}, \href{https://www.adobe.com/acrobat/how-to/delete-pages-from-pdf.html#:~:text=Choose%20%E2%80%9CTools%E2%80%9D%20%3E%20%E2%80%9COrganize,or%20pages%20from%20the%20file.}{Adobe Acrobat} (on all OSs), as well as \href{https://superuser.com/questions/517986/is-it-possible-to-delete-some-pages-of-a-pdf-document}{command line tools}.

\section{Method Details}
% We provide more details of Implicit Decoder, Infinite Depth Query, Gaussian Splatting (GS) head and training strategies.
\subsection{Implicit Decoder.} We provide additional implementation details of the Feed-Forward Network (FFN) and the MLP head in our implicit decoder.

In the FFN, we first expand the input feature dimension by a factor of four, apply a nonlinear activation, and then compress it back to the original dimension.
The MLP head consists of three linear layers with ReLU activations. The input dimension is set to 1024, and the hidden dimension is set to 256.
We use ELU activation after the final layer to avoid vanishing gradient issues during training.

\subsection{Infinite Depth Query}
In the main paper, we illustrate how to obtain the adaptive weight $w_i$ for each pixel $i$.
Here, we describe how to use $w_i$ to select sub-pixel query coordinates.   

Specifically, we normalize $w_i$ into a probability distribution
\begin{equation}
    p_i = \frac{w_i}{\sum_i w_i}.
\end{equation}
Given this discrete distribution $\{p_i\}$, we construct the cumulative distribution function (CDF):
\begin{equation}
    \mathrm{CDF}(k) = \sum_{i=1}^{k} p_i ,
\end{equation}
which is a monotonically increasing function that maps each pixel index $k$ to the total probability mass of all pixels up to $k$.

We then obtain $N$ samples using a uniformly stratified inverse-transform sampling scheme. Specifically, we generate a set of uniformly spaced target values
\begin{equation}
    q_j = \frac{j + 0.5}{N}, \qquad j = 0,\ldots, N-1,
\end{equation}
and for each $q_j$, find the smallest index $k_j$ such that
\begin{equation}
    \mathrm{CDF}(k_j) \ge q_j .
\end{equation}
This yields $N$ pixel indices $\{k_j\}$ whose sampling frequency matches the probability distribution $\{p_i\}$.

For each selected pixel $(u, v)$, we refine the sampling location by adding a random sub-pixel jitter within $[-0.5, 0.5]$ around the pixel center:
\begin{equation}
    (x, y) = \bigl(u + 0.5 + \delta_u,\; v + 0.5 + \delta_v\bigr),
    \quad 
    \delta_u, \delta_v \sim \mathcal{U}(-0.5, 0.5).
\end{equation}
Finally, $(x, y)$ is normalized to match the model's coordinate convention.

\subsection{Gaussian Splatting (GS) Head}
\label{subsec:gshead}
Given the uniform 3D points from Infinite Depth Query, we first enrich each point with color and Plücker ray features extracted from the input image.
These per-point features are then combined with features from the ViT encoder to form point-wise tokens.
Finally, each token is processed through a MLP and fed into multiple independent linear heads to predict Gaussian attributes, including position offsets $o$, color offsets $c$, scales $s$, opacities $\alpha$, and rotations $r$, enabling 3D Gaussian splatting for novel view synthesis.

\subsection{Training Strategies}
We present more details of depth normalization, training InfiniDepth and GS head.
\paragraph{Depth Normalization.}
Before depth normalization, we first convert the ground-truth depth values to logarithmic space to reduce the variance between different scenes. Then, we get the affine-invariant normalized depth using:
\begin{equation}
d_{norm} = \frac{d_{log} - d_{min}}{d_{max} - d_{min}},
\end{equation}
where $d_{log}$ is the logarithmic depth value, and $d_{min}$ and $d_{max}$ are the 2\% and 98\% quantiles of the depth values in the logarithmic space, respectively.
\paragraph{Training InfiniDepth.}
We resize the RGB image but remain the original resolution of the ground-truth depth map, as our implicit depth representation allows us to supervise depth predictions at continuous coordinates.
We construct coordinates-depth pairs on the original ground-truth depth map, and then randomly sample a set of coordinates during training to compute the $l1$ loss. In practice, we sample 100k pairs per image. 
\paragraph{Training GS Head.}
We initialize the ViT encoder with the pretrained InfiniDepth weights and keep it frozen, training only the GS head. The GS head is optimized with a learning rate of $1\times10^{-4}$. Supervision combines an $l1$ reconstruction loss and a perceptual LPIPS loss, encouraging both accurate low-frequency color reproduction and high-frequency structural fidelity in the rendered novel views.

\subsection{Computational Efficiency and Parameter Count}
We provide more analysis on the computational efficiency and parameter count of our model and other baseline models, including  DepthPro~\cite{bochkovskii2024depth},DepthAnythingV2~\cite{yang2024depth2}, MoGe-2~\cite{wang2025moge2}, Marigold~\cite{viola2025marigold}, and PPD~\cite{xu2025pixel}.

As shown in Tab.~\ref{tab:compute_efficiency}, 
the decoder in our model has the lowest parameter count among all compared methods.
The computational efficiency of our model is slower than DepthAnythingV2 and MoGe-2.
However, the convolution decoder used in DepthAnythingV2 and MoGe-2 makes them less effective in capturing fine-grained depth details.
Compared with other methods that also target fine-grained depth estimation, such as DepthPro, Marigold, and PPD, our approach offers better computational efficiency and further surpasses them in the level of detail achieved.

\section{Dataset Details}
\subsection{Synth4K}
\paragraph{Dataset curation.} Synth4K is curated from five different games, including \textit{CyberPunk 2077}, \textit{Marvel's Spider-Man 2}, \textit{Miles Morales}, \textit{Dead Island}, and \textit{Watch Dogs} (Denoted as Synth4K-1, Synth4K-2, Synth4K-3, Synth4K-4, and Synth4K-5, respectively). It contains diverse indoor and outdoor scenes with high-quality graphics and realistic lighting effects. We collect in-game RGB images and corresponding depth maps at a resolution of 3840x2160 (4K) using ReShade, which provides access to the game's rendering pipeline and enables high-quality capture of both color and depth buffers during gameplay.

\paragraph{Implementation of high-frequency mask.} 
To identify high-frequency structures in the depth map $D \in \mathbb{R}^{H\times W}$, 
we compute a geometric energy map that emphasizes local curvature and fine-scale variations.

For a set of smoothing scales $\{s\}$, we first obtain multi-scale filtered depth maps
\begin{equation}
    D_s = 
    \begin{cases}
        \mathrm{GaussianBlur}(D,\, \sigma=s), & s>0,\\[2pt]
        D, & s=0 .
    \end{cases}
\end{equation}

For each scale $s$, we compute the absolute Laplacian response using the 4-neighborhood stencil
\begin{equation}
    \mathcal{L}(D_s) = 
    \left| 
        D_s * 
        \begin{bmatrix}
            0 & 1 & 0 \\
            1 & -4 & 1 \\
            0 & 1 & 0
        \end{bmatrix}
    \right|,
\end{equation}
and aggregate the multi-scale response via a per-pixel maximum:
\begin{equation}
    E(x,y) = \max_{s}\, \mathcal{L}\!\left(D_s\right)(x,y).
\end{equation}
\input{tables/sup/compute.tex}

\input{tables/sup/supervision.tex}
\input{tables/sup/depth_representation.tex}

\input{tables/sup/design.tex}

\begin{figure*}[ht]
    \centering
    \includegraphics[width=0.96\linewidth]{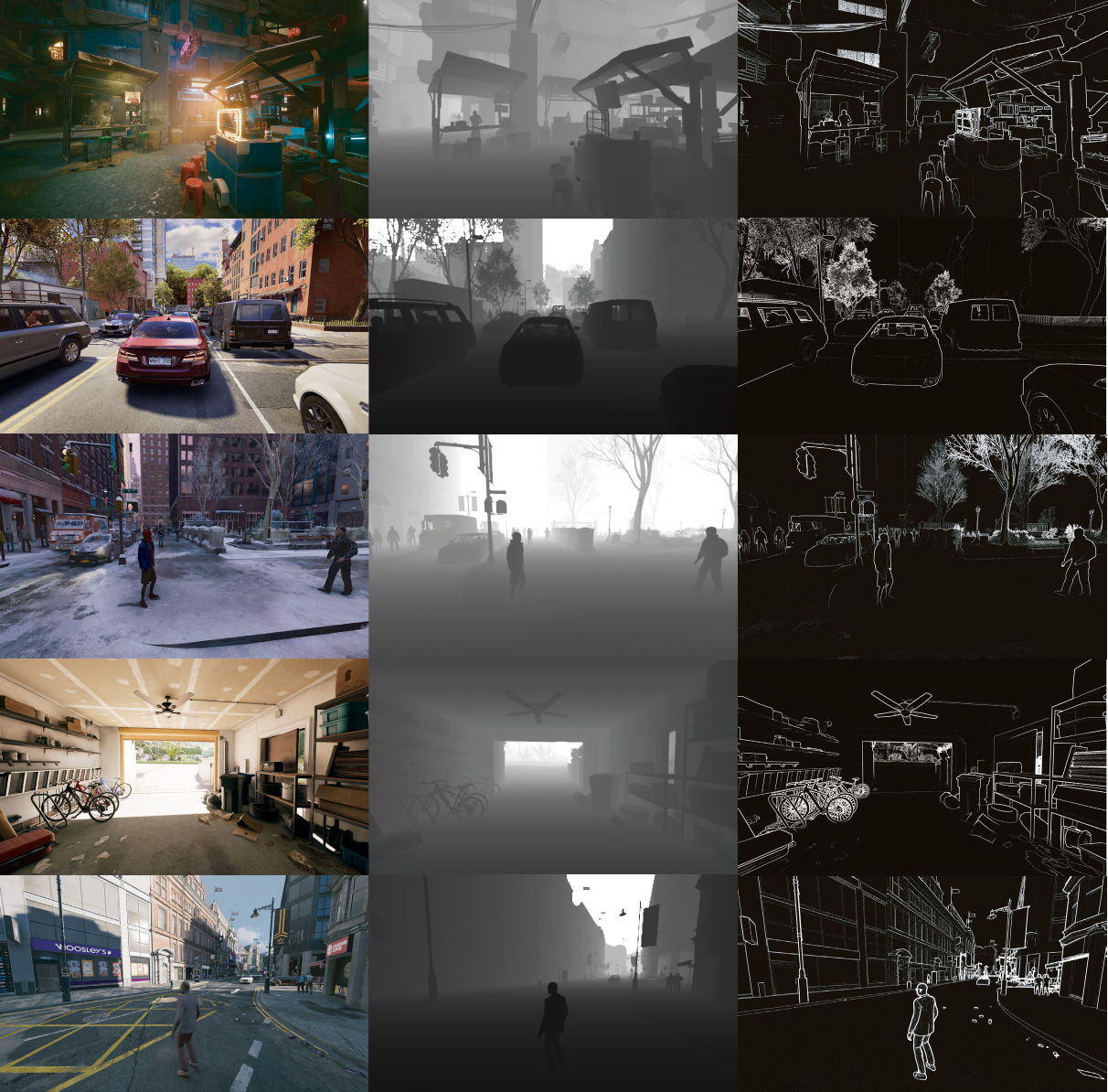}
    \caption{
       \textbf{RGB images, depth maps and high-frequency masks in Synth4K.}
      Each row from top to bottom shows samples from Synth4K's five games: \textit{CyberPunk 2077}, \textit{Marvel's Spider-Man 2}, \textit{Miles Morales}, \textit{Dead Island}, and \textit{Watch Dogs}.
    }
    \label{fig:synth4k}
    \vspace{-3mm}
\end{figure*} 
\begin{figure*}[ht]
    \centering
    \includegraphics[width=0.91\linewidth]{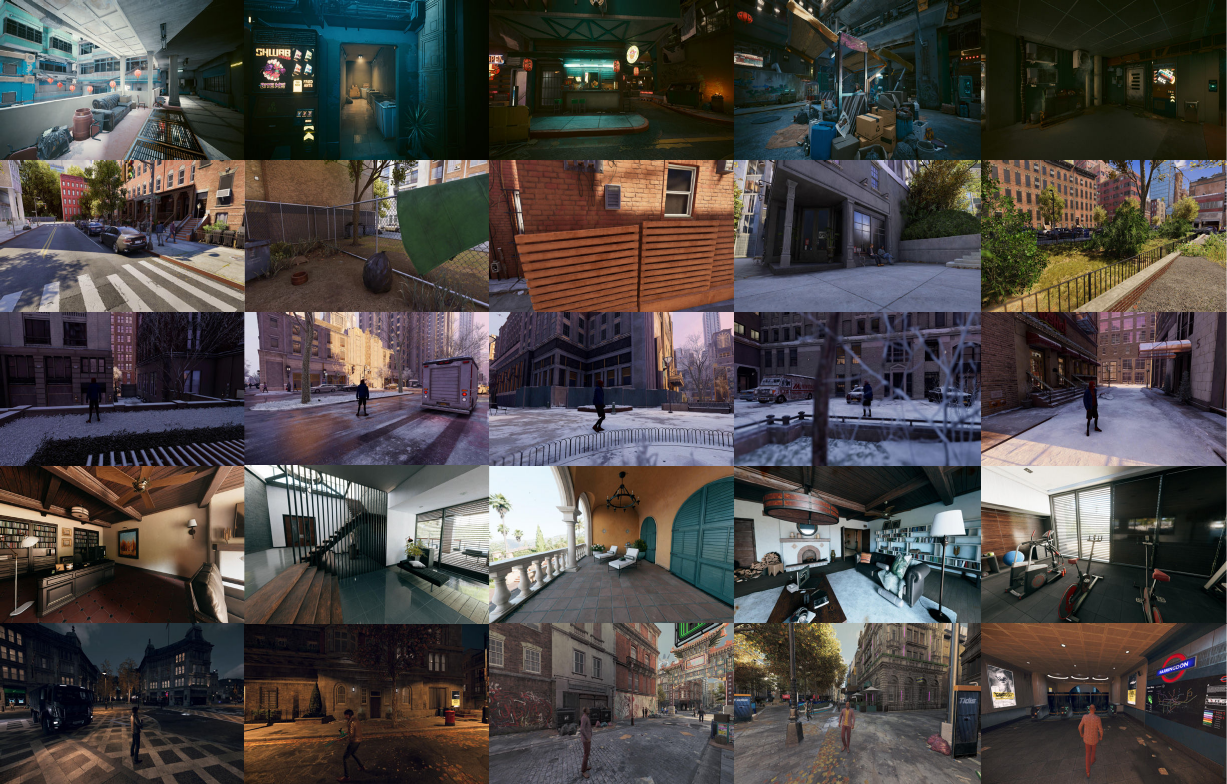}
    \caption{
       \textbf{More RGB images in Synth4K.}
      Each row from top to bottom shows RGB images from Synth4K's five games: \textit{CyberPunk 2077}, \textit{Marvel's Spider-Man 2}, \textit{Miles Morales}, \textit{Dead Island}, and \textit{Watch Dogs}.
    }
    \label{fig:rgb}
    \vspace{-2mm}
\end{figure*}
\begin{figure*}[ht]
    \centering
    \includegraphics[width=0.91\linewidth]{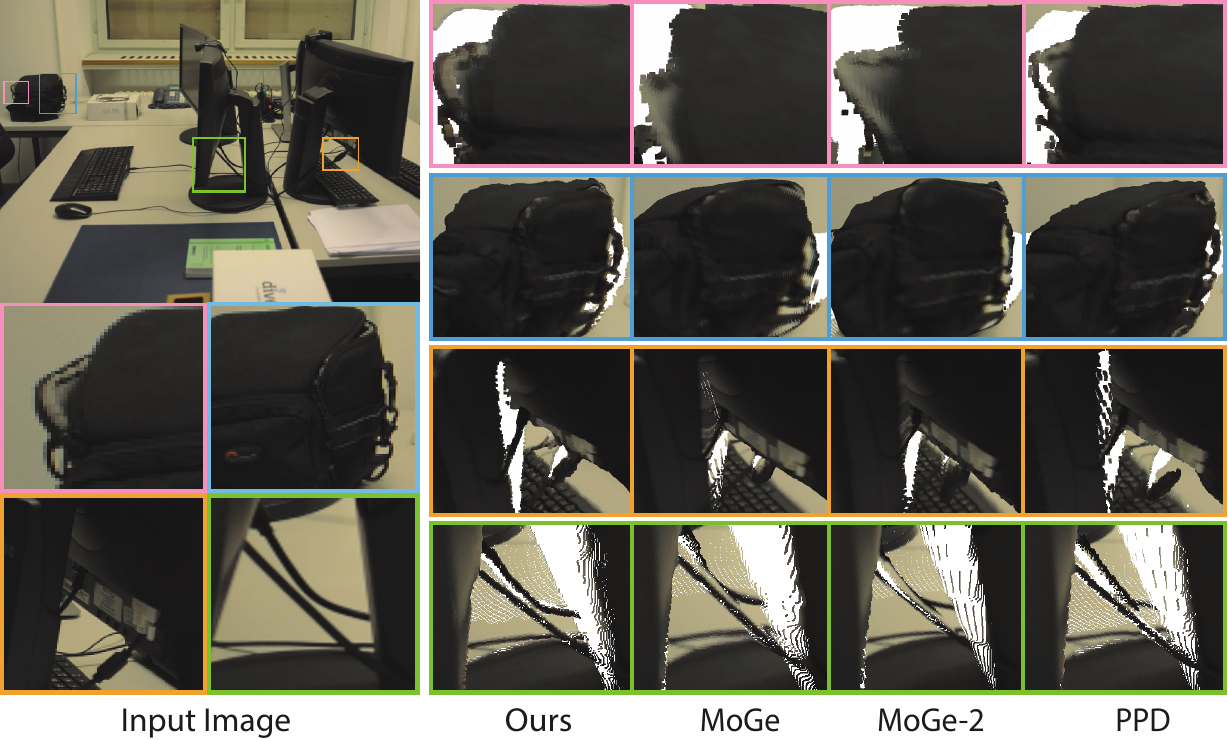}
    \caption{
       \textbf{Point cloud comparisons for relative depth estimation.}
      Each row from top to bottom shows point clouds predicted by our relative depth model and other SOTA models, including MoGe, MoGe-2 and PPD.
    }
    \label{fig:pcd_compare_sup}
    \vspace{-3mm}
\end{figure*} 
\begin{figure*}[ht]
    \centering
    \includegraphics[width=0.91\linewidth]{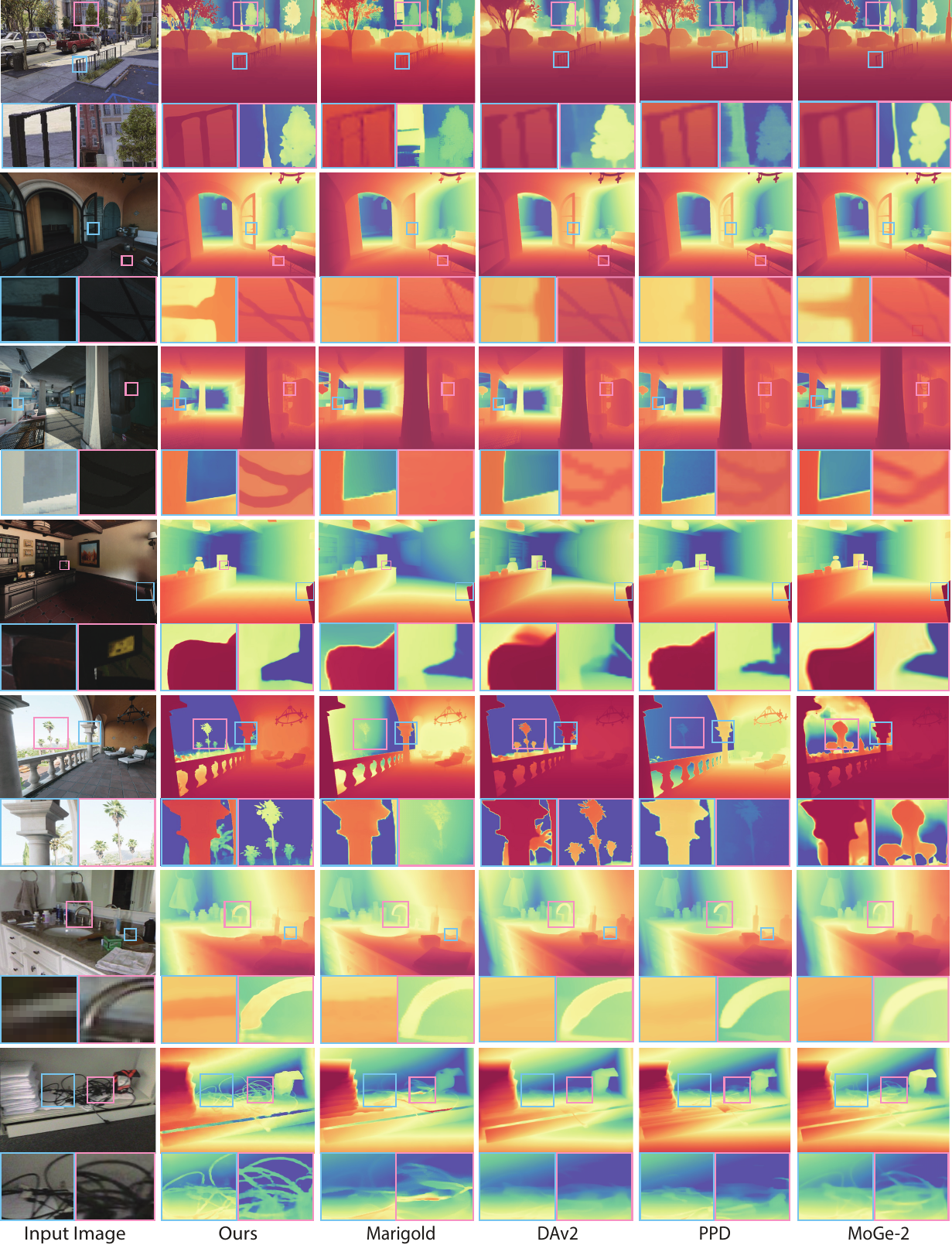}
    \caption{
       \textbf{Depth map comparisons for relative depth estimation.}
      Each row from top to bottom shows depth maps predicted by our relative depth model and other SOTA models, including Marigold, DepthAnythingV2, PPD and MoGe-2.
    }
    \label{fig:depth_compare_sup}
    \vspace{-2mm}
\end{figure*} 
\begin{figure*}[ht]
    \centering
    \includegraphics[width=0.91\linewidth]{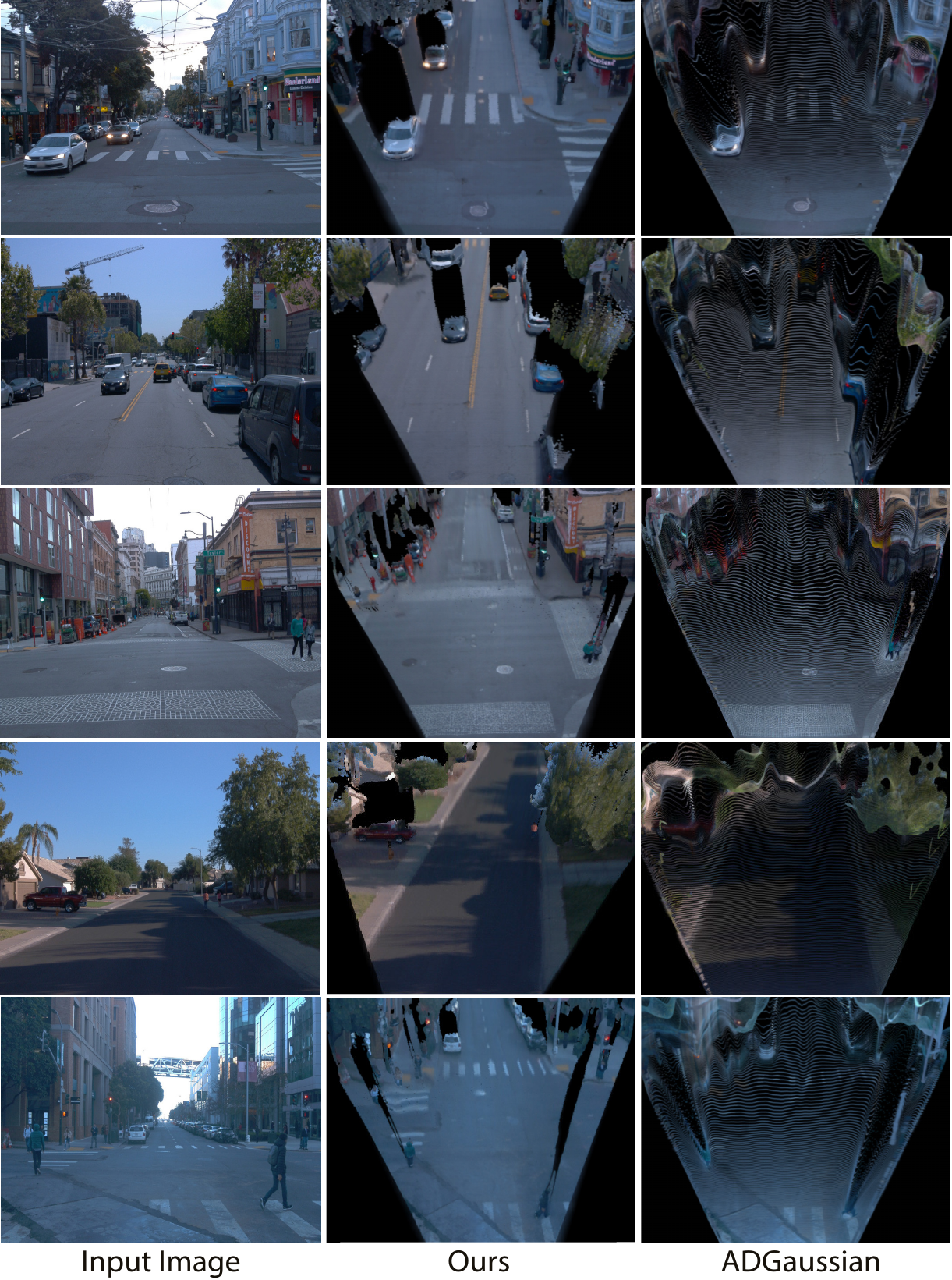}
    \caption{
       \textbf{Single-View Novel View Synthesis (NVS) under large viewpoint shifts.} Each row from top to bottom shows novel view synthesis results from our method and ADGaussian.
    }
    \label{fig:nvs_compare_sup}
    \vspace{-3mm}
\end{figure*} 

To suppress extreme outliers, we normalize $E$ using its $98^\text{th}$ percentile:
\begin{equation}
    \hat{E}(x,y) = 
    \min\!\left( \frac{E(x,y)}{q_{0.98}(E)},\, 1 \right),
\end{equation}
where $q_{0.98}(E)$ denotes the $98\%$ quantile of $E$.

We further apply temperature-based sharpening to control the contrast of the high-frequency response.
Given a temperature parameter $\tau > 0$, we define the sharpened energy as
\begin{equation}
    \tilde{E}(x,y) = \hat{E}(x,y)^{1/\tau}.
\end{equation}
Lower temperature values $(\tau < 1)$ emphasize sharp structures by amplifying large responses,
while higher temperatures $(\tau > 1)$ yield a flatter distribution.

Finally, we compute the sampling probability for each pixel as
\begin{equation}
    p(x,y) = 
    \frac{\tilde{E}(x,y)}{\sum_{x,y} \tilde{E}(x,y)} ,
\end{equation}
and obtain $n$ high-frequency candidate locations by sampling from the discrete distribution 
$\{p(x,y)\}$ using multinomial sampling.

More visualizations about the RGB images, depth maps and high-frequency masks are provided in Fig.~\ref{fig:synth4k} and Fig.~\ref{fig:rgb}.

\subsection{Training Datasets}
Some of our training datasets are introduced in the main paper.
Additionally, we also use the following datasets for training: MatrixCity~\cite{li2023matrixcity}, MVS-Synth~\cite{DeepMVS}, Blendedmvs~\cite{yao2020blendedmvs}, CREStereo~\cite{li2022practical}, FSD~\cite{wen2025foundationstereo}, and DynamicReplica~\cite{karaev2023dynamicstereo}.

\section{Experiments Details}
\subsection{Evaluation Protocols}
We ensure the fair comparison of all methods by using consistent input resolutions and evaluation protocols.

On real-world benchmark, we resize the input image to $504 \times 672$ for all methods, and the output depth maps are evaluated on the same resolution as input, while on Synth4K, we resize the input image to $504 \times 896$ for all methods. The baseline outputs are upsampled to 4K resolution using bilinear interpolation, whereas our method is queried directly at 4K due to the implicit representation.

For the task of relative depth estimation, we align the predicted depth to ground-truth depth using scale-and-shift alignment before evaluation.
For the task of metric depth estimation, we sample 1500 sparse depth points from the ground-truth depth map as additional input for all methods. No alignment is applied during evaluation.
\subsection{Single-View Novel View Synthesis (NVS)}
Single-View Novel View Synthesis (NVS) aims to generate novel views of a scene given a single input image. When the viewpoint changes significantly such as a Bird's-Eye View (BEV), prior methods often produce noticeable artifacts and holes due to incomplete geometry estimation.
We address this challenge by combining our proposed depth representation with a depth query strategy, generating point clouds that uniformly distribute on object surfaces.
Using the Gaussian Splatting (GS) Head described in Sec.~\ref{subsec:gshead}, we can render novel view images from the input RGB image and the uniform point clouds, which produces high-quality results with fewer artifacts and holes.
We train the GS head on a subset of the Waymo~\cite{sun2020scalability} training split and evaluate it on unseen scenes. Qualitative results are shown in Fig.~\ref{fig:nvs_compare_sup}.

\subsection{More Ablation Studies}
\paragraph{Supervision strategies.} We ablate different supervision strategies for training our metric depth model, including sub-pixel supervision and pixel-wise supervision. Sub-pixel supervision refers to using ground-truth depth maps at a higher resolution than the input image during training. This allows us to supervise depth predictions at sub-pixel coordinates within each pixel, which is applied in our full model. Pixel-wise supervision downsamples the ground-truth depth maps to the same resolution of the input image, only providing supervision at the pixel centers. Ablation results in Tab.~\ref{tab:supervision} demonstrate that sub-pixel supervision further improves depth prediction accuracy. It better leverages the inherent property of implicit depth fields to predict depth at continuous coordinates, thereby enhancing the model's ability for fine-grained depth prediction.
\paragraph{Depth representation.} We additionally provide quantitative results of different depth representations for relative depth estimation. Results are shown in Tab.~\ref{tab:depth_representation}. Although the metric accuracy does not improve significantly with neural implicit fields, the visual quality of depth maps is noticeably enhanced, as shown in the main paper.
\paragraph{Design choices of implicit decoder.} Here, we present some different design choices of the feature query module in our implicit decoder, including (1) Coordinate-Offset MLP, (2) Coordinate-Offset MLP (Local Ensemble) and (3) Cross-Attention.
Specifically, for (1), we compute the relative offset between a query coordinate and its nearest grid point, and feed the offset into a shared MLP to learn the input coordinate. We then concat the learned coordinate with the feature of the nearest grid point as the queried feature.
For (2), we compute the relative offsets between a query coordinate and its four surrounding grid points, and then perform similiar operations as (1).
For (3), we use the input coordinate as the $Q$, and the features of its four surrounding grid points as $K$ and $V$s to perform cross-attention for feature aggregation.
We compare the above designs with our default design, which directly uses bilinear interpolation for feature query. Experiments are conducted for metric depth estimation.
As shown in Tab.~\ref{tab:design}, bilinear feature interpolation on feature pyramids achieves the best performance with the least computational cost, while other designs introduce extra parameters and computations but do not lead to performance gains.
We also conduct ablations on different image encoders (DINOv2 vs. DINOv3) for our relative depth model, but observe no significant performance differences.

\section{More Results}
\paragraph{Point Cloud Comparisons.} We additionally provide point cloud comparisons of our relative depth model with other methods, including MoGe, MoGe-2 and PPD. 
As shown in Fig.~\ref{fig:pcd_compare_sup}, our relative depth model demonstrates the strong capability for fine-grained depth estimation. 

\paragraph{Depth Map Comparisons.} 
We provide more depth map comparisons of our relative depth model with 
additional baseline methods, as shown in Fig.~\ref{fig:depth_compare_sup}.

\paragraph{Single-View Novel View Synthesis (NVS) Comparisons.} We present more visual comparisons of single-view NVS results generated by our method and ADGaussian~\cite{song2025adgaussian} to demonstrate the effectiveness of our proposed depth representation and depth query strategy for this task, as shown in Fig.~\ref{fig:nvs_compare_sup}.    

\clearpage

%% file: tables/sup/compute.tex
\begin{table}[ht]
    \centering
    \renewcommand{\arraystretch}{1.2}
    \resizebox{\linewidth}{!}{%
    \begin{tabular}{c|c|c}
        \specialrule{1.2pt}{0pt}{0pt}
        Model & Parameters (M) & Computational Efficiency (s/it) \\
        \hline
        \textbf{Ours} & 15 & 0.16 \\
        DepthPro~\cite{bochkovskii2024depth} & 29 & 0.19 \\
        DepthAnythingv2~\cite{yang2024depth2} & 31 & 0.03 \\
        MoGe-2~\cite{wang2025moge2} & 22 & 0.05 \\
        Marigold~\cite{viola2025marigold} & - & 0.39 \\
        PPD~\cite{xu2025pixel} & - & 1.48 \\
        % Add more models if needed
        \specialrule{1.2pt}{0pt}{0pt}
    \end{tabular}
    }
    \caption{\textbf{Comparison of parameter count and computational efficiency for different decoders.}
    Parameters represent the number of parameters in the decoder, while computational efficiency refers to the inference time required by the entire model to process a single $504 \times 672$ image. We don't report parameters for Marigold and PPD as they are diffusion-based models.
    }
    \label{tab:compute_efficiency}
\end{table}

%% file: tables/sup/supervision.tex
\begin{table*}[t]
    \centering
    \renewcommand{\arraystretch}{1.2}
    \resizebox{\textwidth}{!}{%
    \begin{tabular}{c|c|c|c|c|c|c|c|c|c|c}
    \specialrule{1.2pt}{0pt}{0pt}
    \multirow{2}{*}{Ablation} & Synth4K-1 & Synth4K-2 & Synth4K-3 & Synth4K-4 & Synth4K-5 & KITTI & ETH3D & NYUv2 & ScanNet & DIODE  \\
    \cline{2-11}
    & $\delta_{0.01}$ & $\delta_{0.01}$ & $\delta_{0.01}$ & $\delta_{0.01}$ & $\delta_{0.01}$ & $\delta_{0.01}$ & $\delta_{0.01}$ & $\delta_{0.01}$ & $\delta_{0.01}$ & $\delta_{0.01}$ \\
    \specialrule{1.2pt}{0pt}{0pt}
    \textbf{Sub-Pixel Supervision} & \textbf{72.7} & \textbf{73.5} & \textbf{78.2} & \textbf{81.5} & \textbf{79.4} & \textbf{61.7} & \textbf{93.9} & \textbf{84.7} & \textbf{88.5} & \textbf{97.6}  \\
    Pixel-Wise Supervision & 70.0 & 70.5 & 74.7 & 80.6 & 76.6 & 58.8 & 92.5 & 84.2 & 88.0 & 97.2 \\
    \specialrule{1.2pt}{0pt}{0pt}
    \end{tabular}
    }% end of resizebox
    \caption{\textbf{Quantitative ablations on supervision strategies for metric depth estimation.}}
    \label{tab:supervision}
\end{table*}

%% file: tables/sup/depth_representation.tex
\begin{table*}[ht]
    \centering
    \renewcommand{\arraystretch}{1.2}
    \resizebox{\textwidth}{!}{%
    \begin{tabular}{c|c|c|c|c|c|c|c|c|c|c}
    \specialrule{1.2pt}{0pt}{0pt}
    \multirow{2}{*}{Ablation} & Synth4K-1 & Synth4K-2 & Synth4K-3 & Synth4K-4 & Synth4K-5 & KITTI & ETH3D & NYUv2 & ScanNet & DIODE  \\
    \cline{2-11}
    & $\delta_{0.01}$ & $\delta_{0.01}$ & $\delta_{0.01}$ & $\delta_{0.01}$ & $\delta_{0.01}$ & $\delta_{0.01}$ & $\delta_{0.01}$ & $\delta_{0.01}$ & $\delta_{0.01}$ & $\delta_{0.01}$ \\
    \specialrule{1.2pt}{0pt}{0pt}
    \textbf{Full Model} & \textbf{82.5} & 84.6 & 84.9 & \textbf{93.5} & \textbf{92.5} & \textbf{95.2} & \textbf{98.7} & \textbf{97.3} & \textbf{97.2} & \textbf{96.6}  \\
    w/o Neural Implicit Fields & 82.4 & \textbf{85.3} & \textbf{85.3} & 93.4 & 90.2 & 94.6 & 98.3 & 96.9 & 97.1 & 96.1 \\
    \specialrule{1.2pt}{0pt}{0pt}
    \end{tabular}
    }% end of resizebox
    \caption{\textbf{Quantitative ablations on depth representation for relative depth estimation.} } 
    \label{tab:depth_representation}
\end{table*}

%% file: tables/sup/design.tex
\begin{table}[ht]
    \centering
    \renewcommand{\arraystretch}{1.3}
    \resizebox{\linewidth}{!}{%
    \begin{tabular}{c|c|c|c|c|c}
    \specialrule{1.2pt}{0pt}{0pt}
    \multirow{2}{*}{Ablation}  & KITTI & ETH3D & NYUv2 & ScanNet & DIODE  \\
    \cline{2-6}
    & $\delta_{0.01}$ & $\delta_{0.01}$ & $\delta_{0.01}$ & $\delta_{0.01}$ & $\delta_{0.01}$ \\
    \specialrule{1.2pt}{0pt}{0pt}
    \textbf{Bilinear Feat Interp} & \textbf{61.7} & \textbf{93.9} & \textbf{84.7} & \textbf{88.5} & \textbf{97.6}  \\
    Coordinate-Offset MLP & 59.3  & 90.8  & 80.5  & 81.6  & 96.0  \\
    Coordinate-Offset MLP (Local Ensemble)  & 54.1 & 84.1 & 78.7 & 82.1  & 95.0 \\
    Cross-Attention & 54.8 & 88.2 & 79.7 & 80.7 & 96.2 \\
    \specialrule{1.2pt}{0pt}{0pt}
    \end{tabular}
    }% end of resizebox
    \caption{\textbf{Quantitative ablations on different design choices for metric depth estimation.}
    }
    \label{tab:design}
\end{table}